\documentclass[journal]{IEEEtran}
\ifCLASSINFOpdf
\else
\fi
\usepackage{graphicx}
\usepackage{amsmath}
\usepackage{stfloats}
\usepackage{multirow}
\usepackage{xcolor}
\usepackage{booktabs}
\usepackage{CJKutf8}
\usepackage{pifont}
\usepackage{mathrsfs}
\usepackage{makecell}

\newcommand{\ie}{\emph{i.e.}}

\newcommand{\xmark}{\ding{55}}
\newcommand{\cmark}{\ding{51}}

\begin{document}
%

\title{Learning from Multi-Perception Features for Real-Word Image Super-resolution}

%
%
%

\author{
Axi Niu, Kang Zhang, Trung X. Pham, Pei Wang, Jinqiu Sun, \\ In So Kweon,~\IEEEmembership{Member, ~IEEE}, and Yanning Zhang,~\IEEEmembership{Senior Member,~IEEE}

\thanks{This work was funded in part by the Project of the National Natural Science Foundation of China under Grant 61871328, Natural Science Basic Research Program of Shaanxi under Grant 2021JCW-03, as well as the Joint Funds of the National Natural Science Foundation of China under Grant U19B2037.). (Corresponding author: Jinqiu Sun.)

Axi Niu, Pei Wang, and Yanning Zhang are with the School of Computer Science, Northwestern Polytechnical University, Xi’an, 710072, China, and also with the National Engineering Laboratory for Integrated Aero-Space-Ground-Ocean Big Data Application Technology, Xi’an, 710072, China (email: nax@mail.nwpu.edu.cn,
wangpei23@mail.nwpu.edu.cn, ynzhang@nwpu.edu.cn )


Axi Niu (intern), Kangzhang and Trung X. Pham and In So Kweon are with the School of Electrical Engineering, Korea Advanced Institute of Science and Technology. (email:kangzhang@kaist.ac.kr, trungpx@kaist.ac.kr, iskweon77@kaist.ac.kr)

Jinqiu Sun is with the School of Astronautics, Northwestern Polytechnical University, Xi’an 710072, China (email: sunjinqiu@nwpu.edu.cn)
}
}

%
%

\markboth{Journal of \LaTeX\ Class Files,~Vol.~14, No.~8, August~2015}%
{Shell \MakeLowercase{\textit{et al.}}: Bare Demo of IEEEtran.cls for IEEE Journals}
%



\maketitle


\begin{abstract}

Currently, there are two popular approaches for addressing real-world image super-resolution problems: degradation-estimation-based and blind-based methods. However, degradation-estimation-based methods may be inaccurate in estimating the degradation, making them less applicable to real-world LR images. On the other hand, blind-based methods are often limited by their fixed single perception information, which hinders their ability to handle diverse perceptual characteristics. To overcome this limitation, we propose a novel SR method called MPF-Net that leverages multiple perceptual features of input images. Our method incorporates a Multi-Perception Feature Extraction (MPFE) module to extract diverse perceptual information and a series of newly-designed Cross-Perception Blocks (CPB) to combine this information for effective super-resolution reconstruction. Additionally, we introduce a contrastive regularization term (CR) that improves the model's learning capability by using newly generated HR and LR images as positive and negative samples for ground truth HR. Experimental results on challenging real-world SR datasets demonstrate that our approach significantly outperforms existing state-of-the-art methods in both qualitative and quantitative measures.

\end{abstract}

\begin{IEEEkeywords}
Real-world Image Super-resolution, Multi-Perception Feature Extraction, Cross-Perceived Block, Contrastive Regularization.
\end{IEEEkeywords}

\IEEEpeerreviewmaketitle

\section{Introduction}
\IEEEPARstart{N}{umerous} single image super-resolution (SISR) methods based on CNNs struggle to generalize to real-world low-resolution images due to their dependence on specific degradation scenarios, as the images often undergo arbitrary and complex degradation processes. This results in a significant drop in the performance of CNN-based SISR methods. As a result, researchers are now focusing on developing solutions for the \textit{real-world image super-resolution} (RealSR) problem by estimating degradation using kernel estimate methods in several studies such as blind, KoalaNet, RealSR, and KernelNet~\cite{zuo2019multi,liu2021blind,niu2023cdpmsr,hu2021image,zhang2019improving,wei2018deep,gu2019blind,bell2019blind,ji2020real,liang2021flow,kim2021koalanet,yamac2021kernelnet}.
By injecting the estimated kernel as prior knowledge into the SR model or generating low-resolution and high-resolution image pairs, models can be trained on this additional information. However, these approaches can be highly sensitive and dependent on the accuracy of the estimated kernels, which may differ from the actual kernels. As a result, blind image super-resolution techniques have also emerged as a potential solution to the RealSR problem. This paper also explores blind image super-resolution methods.

Blind image super-resolution is a simple approach for performing super-resolution, where the degradation kernel is unknown and potentially more complex~\cite{yamac2021kernelnet}. Some recent blind super-resolution works have ignored specific degradation processes and patterns, instead designing SR models that directly improve the quality of SR results. These models, such as LP-KPN~\cite{cai2019toward}, CDCnet~\cite{wei2020component}, DDnet~\cite{shi2020ddet}, and ORNet~\cite{li2021learning}, have shown promising results. Blind SR methods typically decompose the super-resolution process into three parts: shallow feature extraction, nonlinear mapping, and reconstruction. They enhance shallow features using a series of single perception convolution operations, with hourglass modules in CDCnet~\cite{wei2020component}, multi-scale dynamic attention in DDnet~\cite{shi2020ddet}, and frequency enhancement units (FEUs)~\cite{li2021learning} in ORNet performing corresponding nonlinear mapping based on the shallow features. These methods have achieved remarkable success in solving real-world SR problems.

Although the previously mentioned methods have shown effectiveness, they all overlook the fact that shallow features from different receptive fields can provide a more adaptive multi-scale reception ability~\cite{li2021perceptual} and that larger receptive fields involve more feature interactions, leading to more refined results and significantly enriching the network's reception scales and perceptual ability~\cite{sun2022shufflemixer}. Therefore, this paper proposes a new approach to solve the real-world image SR problem, which involves learning from multi-perception features to obtain information for SR. Recent studies in other fields have shown that the size and shape of the receptive field determine how the network aggregates local information and considerably affect the overall model performance, such as object detection~\cite{jang2022pooling}, semantic segmentation ~\cite{jang2022pooling}, and image deblur~\cite{li2021perceptual}. In light of this, this paper proposes a novel method for improving RealSR. Specifically, a multi-perception feature extraction unit (MPFE) is designed to obtain diverse perceptual features from the input image. Then, specific cross-perceived blocks (CPBs) can extract more local and global information from the various perceptual features to perform SR. Then, a series of specific cross-perceived blocks (CPBs) is proposed to extract more local and global information from the various perceptual features to perform SR.

Furthermore, recent studies have revealed that using only image reconstruction loss (L1/L2) in low-level tasks may not effectively capture image details and may result in color distortion in the restored images~\cite{wu2021contrastive,wu2021practical}. To address this issue, some methods have incorporated contrastive loss into the optimization of CNN models. For instance, similarly, Wang et al.\cite{wang2021towards} utilized the rest samples in the same mini-batch as negative samples and the output from a teacher network as the positive sample to construct contrastive loss. 
Other approaches~\cite{wang2021unsupervised,zhang2021blind} designed various negative and positive samples, such as patches from the same or different images and feature maps from LR and HR images, to incorporate contrastive loss in their work. Although these methods have demonstrated the effectiveness of contrastive constraint in low-level tasks, it has been pointed out that the dissimilarity between the reconstructed image and the negative samples is too great to contribute to the contrastive loss~\cite{wu2021practical}. To address this limitation, we propose a new contrastive loss by generating multiple hard negative and positive samples, which can ensure that the SR result is pulled closer to the HR image and pushed away from the LR image in the representation space. Overall, the contributions of this work are summarized as follows: 
\begin{itemize}

\item To the best of our knowledge, we are the first to propose an SR architecture for real-world images based on the multi-perception feature construction perspective.

\item Our proposed multi-perception feature extraction unit effectively captures features with different receptive fields, which are crucial for image generation. The cross-perception blocks enable the network to combine the above features, expanding the range of the network's receptive scales and enhancing its perceptual ability.

\item The contrastive regularization term, utilizing specially generated positive and negative samples, encourages the reconstructed image to be closer to the ground truth in the representation space.

\item Extensive experiments on multiple datasets, including RealSR, DRealSR, and RealBlur, demonstrate that our MPF-Net outperforms state-of-the-art methods in both quantitative and qualitative evaluations, highlighting the effectiveness of our proposed architecture.

\end{itemize}

\vspace{-10pt}
\section{Related Work}

Single Image super-resolution (SISR) is a classic ill-posed inverse problem, which has attracted much attention because of its wide range of applications~\cite{pan2020unsupervised,pan2022ml,pan2022labeling,lei2020deep,hu2019channel}. Therefore, more and more related studies have come out. In this section,  we focus on introducing  kernel-estimated super-resolution methods and blind super-resolution methods.

\vspace{-10pt}
\subsection{Kernel-estimated Super-Resolution Methods}
As described in~\cite{deng2015single,yamac2021kernelnet,zhang2020deep,huang2020unfolding,wang2021unsupervised}, the degradation of an HR image can be seen as the following process:
\begin{equation}
    y=(x\otimes k)\downarrow _{ds} + n,
    \label{eq:generate_lr}
\end{equation}
where $x$ and $y$ represent the HR image and LR image, respectively. $\otimes $ denotes a two-dimensional convolution operated on $x$ with blur kernel $k$. $n$ represents Additive White Gaussian Noise (AWGN), and $\downarrow _{ds}$ is the standard downsampler. SISR refers to the process of recovering $x$ from $y$. 

In recent years, several kernel-based super-resolution (SR) methods have gained popularity in the research community. Among these methods, KernelGAN~\cite{bell2019blind} has gained significant popularity for solving real-world SR problems using kernel estimation. This success has inspired the development of several other kernel-based SR methods. For example, IKC~\cite{gu2019blind} uses an iterative kernel correction method to estimate blur kernels and employs spatial feature transform layers to process these kernels efficiently. RealSR~\cite{ji2020real} employs a degradation framework for real-world images by estimating various blur kernels and real noise distributions and injecting them into its network. FKP~\cite{liang2021flow} generates reasonable kernel initialization by learning an invertible mapping between the anisotropic Gaussian kernel distribution and a tractable latent distribution. KOALAnet~\cite{kim2021koalanet}, on the other hand, jointly learns spatially-variant degradation and restoration kernels to adapt to the spatially variant blur characteristics in real images. Finally, KernelNet~\cite{yamac2021kernelnet} proposes a modular and interpretable neural network  for blind SR kernel estimation. These methods rely on estimating a good degradation kernel and simulating the degradation process, and thus their performance heavily depends on the accuracy of the estimated degenerate kernel.

\vspace{-10pt}
\subsection{Blind Super-Resolution Methods}

Blind image super-resolution is a popular approach for addressing real-world SR problems, where the specific degradation process and pattern are unknown, and models are directly designed to improve SR results. For instance, 
LP-KPN~\cite{cai2019toward}, and CDCnet~\cite{wei2020component} propose various methods for predicting non-uniform degradation kernels using Laplacian pyramid-based kernel prediction networks, per-pixel kernel learning, and Component-Attentive Blocks (CABs), respectively. DDnet~\cite{shi2020ddet} introduces a dual-path dynamic enhancement network that utilizes multiple dynamic kernels with various sizes for information aggregation to capture more multi-scale information effectively. Then, MS2Net~\cite{niu2022ms2net} designs a dual-branch architecture to super-resolve motion-blur and low-resolution images, which have achieved superior performance on public datasets.  Moreover, other SR methods, other SR methods such as~\cite{huang2020fast,chang2021two}, propose  new optimization methods for current blind image super-resolution. The frequency enhancement unit (FEU) in~\cite{li2021learning} performs nonlinear mapping based on shallow features from a single perception convolution operation. These methods have demonstrated significant success in solving real-world SR problems.

However, they typically rely on single-perspective information and overlook the fact that different perspective features may provide more diverse and colorful information to support image super-resolution~\cite{lim2017enhanced,zhang2018image,wang2018esrgan,mei2021image,liang2021swinir,niu2023gran}. As shown in Tab.~\ref{tab:mpfe_cpb}, we have demonstrated that multi-perspective features perform well than single ones.   

\subsection{Contrastive Learning (CL)}
Contrastive Learning has gained great popularity for self-supervised representation learning~\cite{zhu2022self,fan2022partial,pham2022pros,niu2022fast,xue2022slimseg,pham2022self}. It aims to bring anchor samples close to positive samples and simultaneously while pushing them away from negative ones by mapping them to a high-dimensional representation space~\cite{pham2022pros,chen2022consistent,pham2021self}. Recently, CL has been applied to low-level tasks, such as image dehazing~\cite{wu2021contrastive}, degradation representation learning for super-resolution~\cite{wang2021unsupervised}, and even for real image super-resolution using Criteria Comparative Learning~\cite{shi2022criteria}, which defines the contrastive loss directly on criteria instead of image patches. However, previous attempts at applying contrastive learning to image super-resolution (SR) have used HR images as positive and LR images as negative samples. In this paper, we propose a novel contrastive learning approach based on specifically generated positive and negative samples that can better push the restored image closer to the clear image in the representation space.

\begin{figure*}[!ht]
\centering	
\includegraphics[width=18 cm]{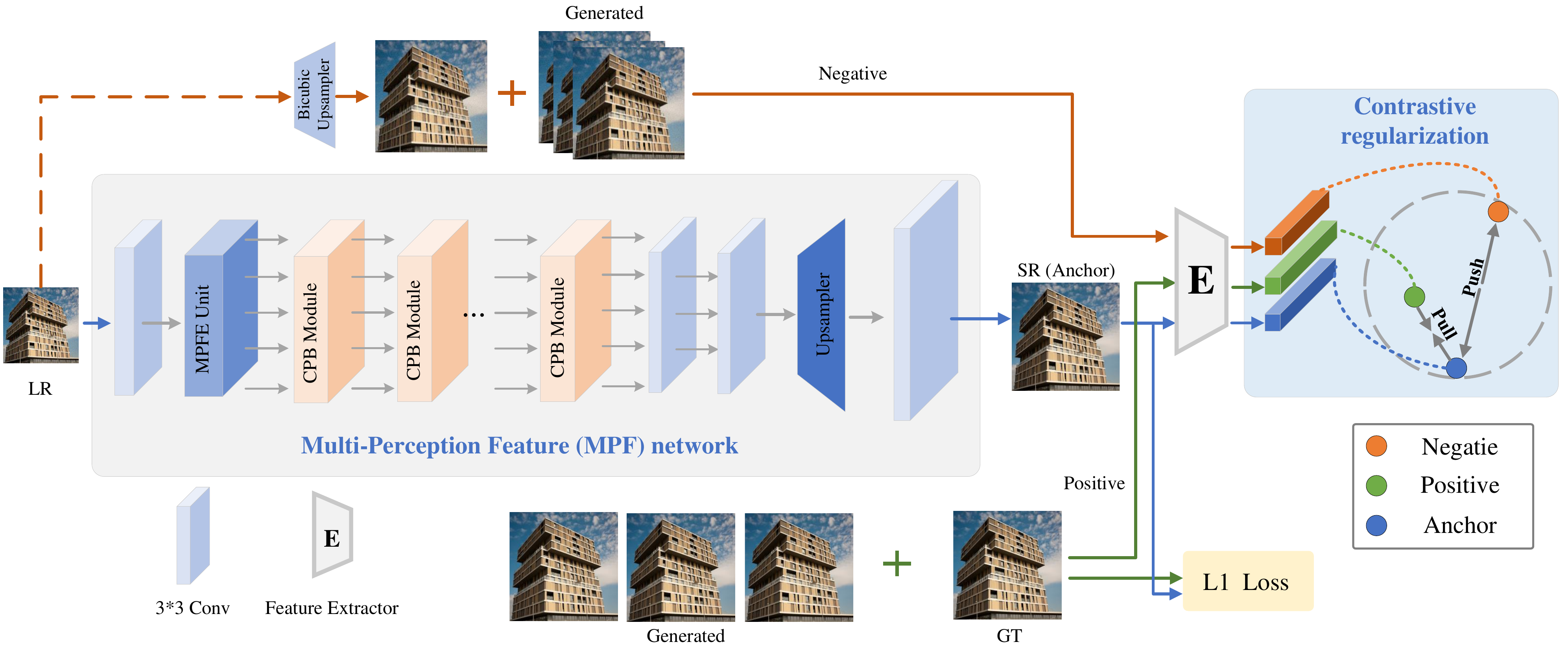}
	\caption{
	The architecture of the proposed MPF-Net. It comprises two components: the multi-perception feature (MPF) super-resolution network and the contrastive regularization item (CR). We jointly minimize the L1-based reconstruction loss and contrastive regularization to pull the super-resolved image better (\ie anchor) to the clear (\ie positive) image and push the restored image to the hazy (\ie negative) image.
 }
	\label{fig:overview}
 \vspace{-10pt}
\end{figure*}

\section{Methodology}
This section begins with an introduction to our multi-perception feature network (MPF-Net), followed by the presentation of the multi-perception feature extraction unit (MPFE), which is designed to extract diverse perceptual features from input images, and the cross-perception block (CPB), which enhances the information in each perceptual domain. Finally, we apply contrastive regularization (CR) as a universal regularization technique to our MPF-Net.

\subsection{The overview of our Multi-Perception Feature Network}

The main goal of our work is to develop a universal and effective approach for enhancing the resolution of real-world images. Our framework, illustrated in Fig.\ref{fig:overview}, involves a comprehensive workflow starting from a low-resolution image $I_{LR}$ and aiming to produce a sharp and high-resolution image $I_{SR}$. To improve the super-resolution performance on real-world images, we propose a multi-perception feature network that extracts features from various perspectives. Our approach includes two technical innovations: the Multi-Perception Feature Extraction (MPFE) unit and the Cross-Perception Block (CPB), as depicted in Fig.\ref{fig:mpfe} and Fig.~\ref{fig:cpb}. The MPFE unit uses different convolutional operations to extract features of diverse perspectives, while the CPB performs non-linear mapping of these features to obtain the final information for image reconstruction. The overall processing flow can be represented by the following equation:

\begin{equation}
\begin{aligned}
    & F_{MPFE} =H_{MPFE}(I_{LR}),\\
    & F_{CPB}=H_{n}^{CPB}(...(H_{2}^{CPB}(H_{1}^{CPB}(F_{MPFE})))...),\\
    & I_{SR}=conv(H_{up}(conv(conv(F_{CPB})))),
\end{aligned}
\end{equation}
where $H_{MPFE}$, $H_{n}^{CPB}$ and $H_{up}$ correspond to the Multi-Perception Feature Extraction (MPFE) unit, Cross-Perception Block (CPB), and pixel shuffle operation, respectively. The outputs of the MPFE and CPBs  are represented by $F_{MPFE}$ and $F_{CPB}$, respectively, with $n$ being the number of CPBs. In our approach, we have set $n$ to 10.

Additionally, our single image super-resolution (SISR) method is designed to reconstruct low-resolution images using two distinct losses: an image reconstruction loss and a contrastive regularization term. These losses can be mathematically expressed as follows:
\begin{equation}
\begin{aligned}
    L_{\text{total}}&=L_{\text{recon}} + \beta L_{\text{CR}}\\
             &=arg\min_{\theta } \left \| I_{SR} -\phi(I_{LR},\theta )  \right \| + \beta \rho (\phi(I_{LR},\theta )),
\end{aligned}
\label{eq:loss_total_1}
\end{equation}
where $I_{LR}$ denotes a low-resolution image, $I_{SR}$ represents the corresponding super-resolved image, and $\phi(\cdot,\theta)$ is our SR network with $\theta$ as its parameters. The term $\left | I_{SR} -\phi(I_{LR},\theta ) \right |$ is the image reconstruction loss, which is typically formulated using an L1/L2 norm-based loss function. The contrastive regularization term, denoted by $\rho(.)$, is applied to produce a natural and smooth super-resolved image. Lastly, $\beta$ is a penalty parameter used to balance the weight between the reconstruction loss and the contrastive regularization term.

 \vspace{-10pt}
\subsection{Multi-Perception Feature Extraction Unit (MPFE)}
\label{sec:MPEF}

One of the main challenges in real-world SR is the significant variation in the degrees and scales of degradation patterns. Traditional real-world SR methods employ fixed and inflexible single-perception features, which limit the models' reception variety and perceptual ability. Although non-local modules proposed by~\cite{wang2018non,li2021perceptual,mei2021image,zhou2023memory} can be a potential solution, they require substantial computation and memory expenses, as shown in Tab.\ref{tab:param_flop_time}. Additionally, compared to our multi-perceptive feature, the non-local module needs more weight maps to obtain features. Inspired by\cite{li2021perceptual,jang2022pooling,sun2022shufflemixer}, which demonstrated that different receptive fields' convolution operations could affect how the network extracts information and its performance, we investigated whether multi-perception can help SR tasks.

\begin{figure}[!ht]
	\centering	
	\includegraphics[width=7.5 cm]{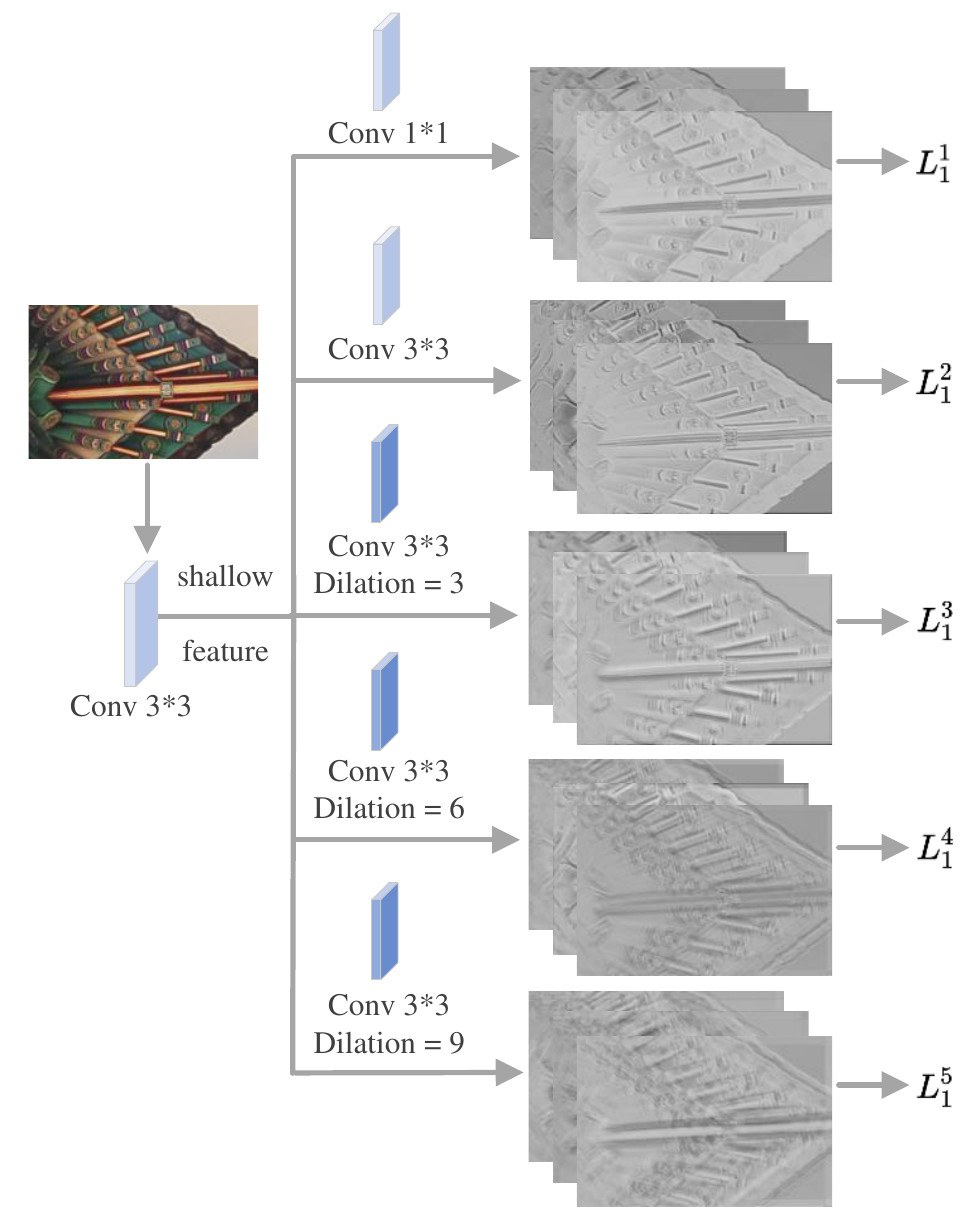}
	\caption{A real example that the MPFE unit extracts the information from reception  files of 3 scales with different dilation and 2 vanilla convolutions.}
	\label{fig:mpfe}
 \vspace{-15pt}
\end{figure}

In contrast to previous SR methods, we propose a novel Multi-Perception Feature Extraction Unit (MPFE) that obtains multi-perception features using various convolution operations after achieving shallow features in the first convolution layer. Our MPFE, as shown in Fig.\ref{fig:mpfe}, consists of one vanilla $1\times1$ conv, one vanilla $3\times3$ conv, and three different dilated convolutions. The five features obtained from the MPFE with different receptive fields have a reasonable perceptual range, from small to large and from fixed to flexible, as shown in Fig.\ref{fig:mpfe}. This enriches the network's reception scales and perceptual ability.

\subsection{Cross-Perception Block (CPB)}
\begin{figure}[!ht]
	\centering	
	\includegraphics[width=5.5 cm]{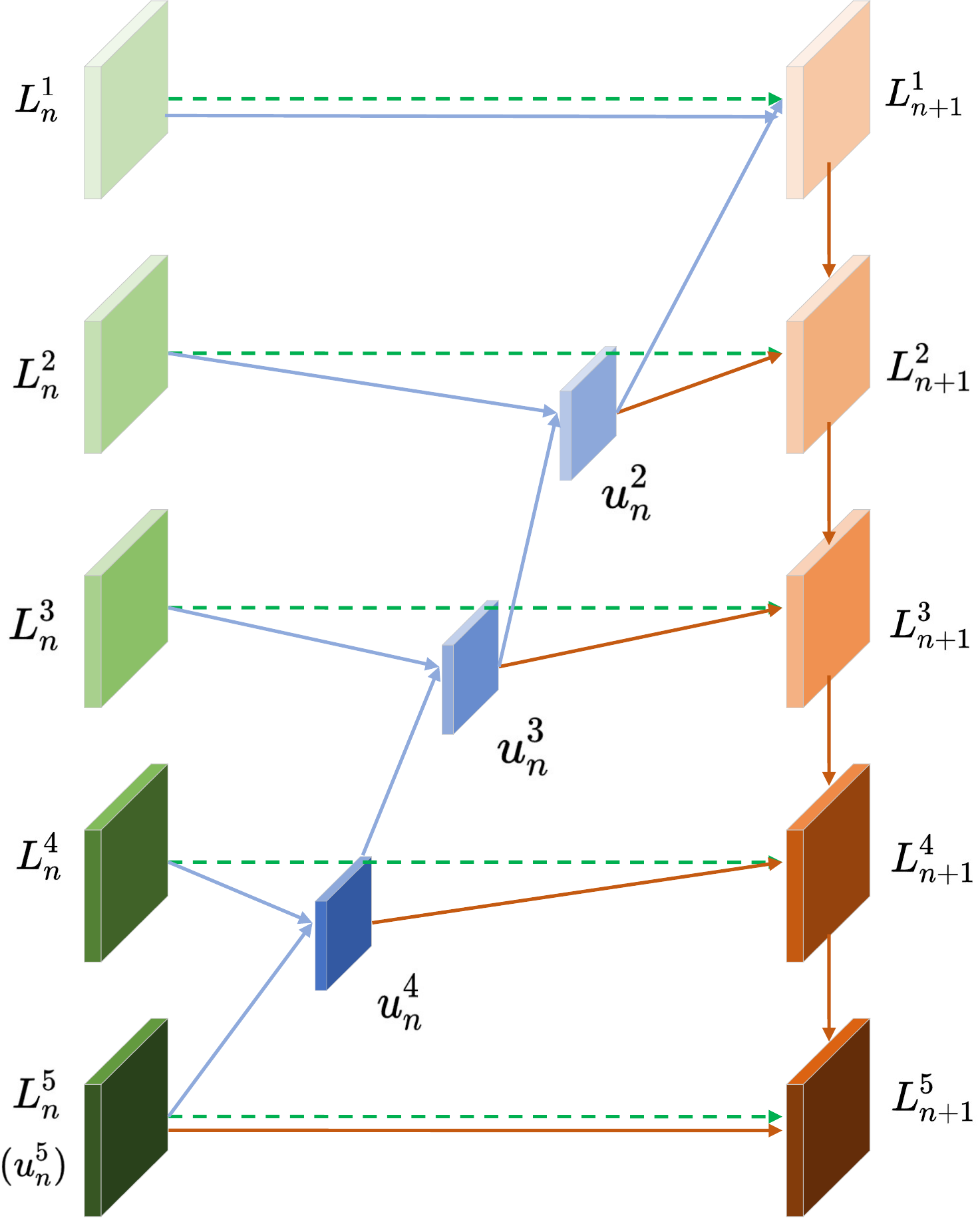}
	\caption{The overview of our CPB block. Inspired by EfficientDet~\cite{tan2020efficientdet}, we design Our Cross-Perception Block (CPB) to fuse multi-reception features by incorporating an up-to-down and down-to-up pathway. The down-to-up pathway (blue line) involves concatenating multi-reception features from the MPFE unit, while the up-to-down pathway (red line) concatenates the downsampled features. The two pathways are then fused using an add operation (green dotted line).
    }
	\label{fig:cpb}
\end{figure}

We have adopted the highly effective BiFPN block from EfficientDet~\cite{tan2020efficientdet} as the basic block in our proposed Multi-Perception Feature Network (MPE-Net), as it has been successful in object detection tasks. However, unlike the BiFPN block designed for cross-scale features, our Cross-Perception Block (CPB) aims to enhance information in each perceptual domain. The CPB, like the BiFPN block, provides abundant adaptive multi-perception reception ability. It operates on the five different receptive features through a down-top pathway to obtain intermediate outputs, which are then operated on by a top-down pathway, resulting in fully integrated features that contain both smooth content and rich details. CPBs, with their perceptual variousness, can perceive and adapt to various degradation patterns with large distribution scales. Using CPB multiple times can significantly broaden the network's reception scales and perceptual ability, which is beneficial for removing degradation.

Specifically, as shown in Fig.~\ref{fig:cpb}, the $n$-th Cross-Perception Block (CPB) takes the input $L_{n}^{i}$ and produces an output $L_{n+1}^{i}$, where $n$ ranges from 1 to 10, and $i$ ranges from 1 to 5. The computation of the $n$-th CPB can be expressed as follows:

\begin{equation}
    \begin{aligned}
        u_{n}^{5} &= L_{n}^{5} \\
        u_{n}^{i} &= \text{Block}(\text{concat}(L_{n}^{i},u_{n}^{i+1})), \quad i = {4,3,2,1}
    \end{aligned}
\end{equation}

\begin{equation}
    \begin{aligned}
        d_{n}^{1} &= u_{n}^{1} \\
        d_{n}^{i} &= \text{Block}(\text{concat}(u_{n}^{i},d_{n}^{i-1})), \quad i = {2,3,4,5}
    \end{aligned}
\end{equation}

\begin{equation}
    \begin{aligned}
        L_{n+1}^{i} = L_{n}^{i} + d_{n}^{i}, \quad  i = {1,2,3,4,5}
    \end{aligned}
\end{equation}
where $L_{n}^{i}, i=1,2,...,5$ means the input of the $n$-th CBP block. When $i=1$, it represents the output of the MPFE unit. The intermediate output of the Block operation on the output of the down-to-top pathway in CPB is denoted as $u_n^i$, where $i$ takes values from 1 to 5. The intermediate output of the Block operation on the output of the top-to-down pathway is denoted as $d_n^i$, where $i$ takes values from 2 to 5. In this paper, we use 10 CPB blocks,~\ie $n=10$. The Block operation consists of two 3$\times$3 convolution tails with a SELU activation function and a squeeze-and-excitation~\cite{hu2018squeeze} layer.

\subsection{Contrastive Regularization}
\label{sec:CR}

Taking inspiration from contrastive learning~\cite{wu2021practical}, our proposed contrastive regularization (CR) aims to bring the anchor sample closer to positive samples and push it farther away from negative samples in latent space. Our approach is designed to generate better super-resolved images. Unlike previous attempts at applying contrastive learning to low-level tasks~\cite{wu2021contrastive,wang2021towards}, our CR is constructed by generating distinct positive and negative samples. In the subsequent sections, we will elaborate on how we generate positive and negative samples and how we calculate our CR.

\textbf{Positive Sample Generation.}
To address the one-to-many problem of SISR, where one LR image corresponds to multiple HR images, we generate additional positive samples instead of solely relying on ground truth images. Similar to~\cite{wu2021practical} and~\cite{wang2021towards}, we utilize high-pass filtering and a teacher network to generate positive samples. Specifically, for the $i$-th LR image, we generate its positive set through the following steps:

\begin{equation}
   P_{i} =\left \{ P_{j}|P_{j}=H(I_{i}^{HR} )  \right \} _{j=1}^{K_{P} } + \mathscr{U}(I_{i}^{LR}) + I_{GT},
\end{equation}
where the $H$ presents a random high-pass filter, and the $K_{P}$ is the number of positive samples generated by high-pass filters. We take a pre-trained SR $\mathscr{U}$ as the teacher network to generate an additional positive sample. $I_{GT}$ and $P_{i}$ mean the ground truth and all the positive samples, respectively. Here, we set $K_{P}=3$. Therefore, the total number of positive samples used in our paper is 5.

\textbf{Negative Sample Generation.} The negative samples in existing low-level image restoration works~\cite{wu2021contrastive,wang2021towards} often include the LR image itself or other LR images in the same batch size, which have been found to be dissimilar to the reconstructed images and easily distinguishable~\cite{wu2021practical}. To address this issue, we draw inspiration from~\cite{wu2021practical} and generate negative samples by introducing slight random Gaussian blur and noise to the ground truth image. For the $i$-th image, the specific process for generating its negative set is as follows:

\begin{equation}
   N_{i} =\left \{ N_{j}|N_{j}=B(I_{i}^{HR} )+n  \right \} _{j=1}^{K_{N} } + I_{i}^{LR},
\end{equation}
where $B$ and $n$ denote random Gaussian blur and Gaussian noise, respectively. $K_N$ is the number of negative samples generated, and $I_{i}^{HR}$ represents the ground truth image. Here, we set $K_N$ to be 4, so the total number of negative sample sets in our paper is 5.

\begin{table*}
\begin{center}
\caption{ Quantitative results on the RealSR dataset. We compare our MPF-Net to the general SISR methods, including Bicubic, EDSR, RCAN, ESRGAN, NLSA, kernel-based SISR methods, including IKC, DAN, DASR, and RealSR methods, including 
LP-KPN, CDC, OR-Net, MS2Net, and Cria-CL. We use PSNR, SSIM, and LPIPS as evaluation metrics.
}\label{tab:realsr}
\resizebox{0.9\hsize}{!}{
\begin{tabular}{c|c|c|ccc|c|ccc|c|ccc}
\toprule
\hline
\multirow{2}{*}{Method} & \multirow{2}{*}{Category} & \multirow{2}{*}{Scale} & \multicolumn{3}{c|}{RealSR}      & \multirow{2}{*}{Scale} & \multicolumn{3}{c|}{RealSR}      & \multirow{2}{*}{Scale} & \multicolumn{3}{c}{RealSR}      \\ \cline{4-6} \cline{8-10} \cline{12-14}       &        &        & \multicolumn{1}{c}{PSNR} & SSIM  &LPIPS  &         & \multicolumn{1}{c}{PSNR} & SSIM  &LPIPS &            & \multicolumn{1}{c|}{PSNR} & SSIM  &\multicolumn{1}{c}{LPIPS} \\ \hline
Bicubic                 & \multirow{5}{*}{\makecell[c]{vanilla \\ SISR}}     & \multirow{5}{*}{X2}    & 31.67  & 0.8870 &0.2227   & \multirow{5}{*}{X3}      & 28.61   & 0.8008 & 0.3891     & \multirow{5}{*}{X4}      & 27.24      & 0.7635    &0.4764        \\ 
EDSR~\cite{lim2017enhanced}      &      &     & 33.88    & 0.9195 &0.1453  &    & 30.86  & 0.8667 &0.2192  &   & 29.09  & 0.8270   &0.2779                \\ 
RCAN~\cite{zhang2018image}       &        &      & 33.83     & 0.9226 &0.1472  &        & 30.90        & 0.8642 &0.2254   &        & 29.21       & 0.8237  & 0.2868           \\ 
ESRGAN~\cite{wang2018esrgan}                    &   &          &33.80     &0.9224  &0.1463   &     & 30.72     &0.8663 &0.2194  &        & 29.15     &0.8263 & 0.2793   \\  
NLSA~\cite{mei2021image}        &      &       & 33.93       &0.9274                             &0.1303         &       &30.93   &0.8711  &0.2120   &    &    29.23   &0.8281   &0.2651                      \\ 
\midrule
IKC~\cite{gu2019blind}                     & \multirow{3}{*}{\makecell[c]{kernel-based \\ SISR} }     & \multirow{3}{*}{X2}    & 33.24       &0.9186  &0.1342  & \multirow{3}{*}{X3}  & 23.48    & 0.7723  &0.2382   & \multirow{3}{*}{X4}      & 16.81         & 0.5277       &0.3824      \\ 
DAN~\cite{huang2020unfolding}               &       &         & 32.29      &0.8992   &0.1726   &             &  29.16     & 0.8277      &0.3199   &       & 27.80       &  0.7882      &0.4114     \\ 
DASR~\cite{wang2021unsupervised}       &         &         &32.24     &0.8979      &0.1814  &   & 29.14      &0.8274      &0.3214   &    &     27.79      & 0.7874    &0.4076   \\  \midrule
LP-KPN~\cite{cai2019toward}        & \multirow{6}{*}{\makecell[c]{real-world \\ SISR}}     &\multirow{6}{*}{X2}       & 33.90     &0.9265  &-   & \multirow{6}{*}{X3}     & 30.60   & 0.8675   &-   &\multirow{6}{*}{X4}    & 29.05   & 0.8335 &-   \\ 
CDC~\cite{wei2020component}     &       &     & 33.96   & 0.9245 &0.1418    &    & 30.99   & 0.8686  &0.2145     &   & 29.24   & 0.8265     & 0.2781    \\ 

OR-Net~\cite{li2021learning}                  &                           &                        & 34.08                      & 0.9281 &-   &           & -                     & - &-    &           & -                     & -     &-              \\
MS2Net~\cite{niu2022ms2net}          &             &         &33.83   &0.9226   &0.1423 &           & -   & - & - &            & 27.33     & 0.7869    &  0.3565      \\ 
Cria-CL~\cite{shi2022criteria}                  &              &                        &-                      &-     &- &           & -                     & - &-  &            & 25.83         & 0.7324    &0.4121     \\ 
\textbf{MPF-Net (Ours)}              &           &         &\textbf{34.25}   &\textbf{0.9302}  &\textbf{ 0.1409 }  &   &\textbf{31.11}    &\textbf{0.8701}  &\textbf{0.2113}   &    &\textbf{29.50}      &\textbf{0.8288}  & \textbf{0.2664}    \\ \hline
\bottomrule
\end{tabular}}
\end{center}
 \vspace{-10pt}
\end{table*}

\begin{table*}
\begin{center}
\caption{ Quantitative results on the DRealSR dataset. We compare our MPF-Net to the general SISR methods, including Bicubic,  EDSR, RCAN, ESRGAN, NLSA, kernel-based SISR methods, including IKC, DAN, DASR, and RealSR methods, including 
, LP-KPN and CDC, OR-Net, MS2Net, and Cria-CL. We use PSNR, SSIM, and LPIPS as evaluation metrics.}
\label{tab:drealsr}
\resizebox{0.9\hsize}{!}{
\begin{tabular}{c|c|c|ccc|c|ccc|c|ccc}
\toprule
\hline
\multirow{2}{*}{Method} & \multirow{2}{*}{Category} & \multirow{2}{*}{Scale} & \multicolumn{3}{c|}{DRealSR}      & \multirow{2}{*}{Scale} & \multicolumn{3}{c|}{DRealSR}      & \multirow{2}{*}{Scale} & \multicolumn{3}{c}{DRealSR}  \\
\cline{4-6} \cline{8-10} \cline{12-14}   
                        &                           &                        & \multicolumn{1}{c}{PSNR} & SSIM & LPIPS &                        & \multicolumn{1}{c}{PSNR} & SSIM & LPIPS &                        & \multicolumn{1}{c}{PSNR} & SSIM & \multicolumn{1}{c}{LPIPS} \\ 
\hline
Bicubic                 & \multirow{5}{*}{\makecell[c]{vanilla \\ SISR}}     & \multirow{5}{*}{X2}    & 32.67       & 0.8771 & 0.2011      & \multirow{5}{*}{X3}      & 31.50      & 0.8352 & 0.3615      & \multirow{5}{*}{X4}   & 30.56       & 0.8200 & 0.4376     \\ 
EDSR~\cite{lim2017enhanced}      &             &        & 34.24   & 0.9083 &  0.1555 &                      & 32.93     & 0.8763 & 0.2413 &      & 32.03        & 0.8551    &0.3071  \\ 
RCAN~\cite{zhang2018image}          &      &       & 34.34  &   0.9080 & 0.1583 &     & 33.03           & 0.8760 & 0.2413 &       & 31.85    & 0.8571 &0.3054       \\ 
ESRGAN~\cite{cheng2019encoder}          &             &   & 33.89      & 0.9061 &0.1556      &     & 32.39     &0.8733    &0.2432  &       &31.92     &0.8570  &0.3081 \\  
NLSA~\cite{mei2021image}      &        &          & 34.01       &0.9102     &0.1514   &                        &32.46    &0.8729   &0.2411  &  &32.11     &0.8601   &0.3011  \\  
\midrule
IKC~\cite{gu2019blind}          & \multirow{3}{*}{ \makecell[c]{ kernel-based \\ SISR}}     & \multirow{3}{*}{X2}    & 34.14    &0.9126  &\textbf{0.1131}     & \multirow{3}{*}{X3}      &27.29         &0.8426    &\textbf{0.2291}   & \multirow{3}{*}{X4}      &21.56      &0.6841  &0.3553 \\ 
DAN~\cite{huang2020unfolding}    &        &         &32.51      &0.9033   &0.1729    &                        &31.54            & 0.8421      &0.3217  &      &30.59          & 0.8201  &0.4111 \\ 
DASR~\cite{wang2021unsupervised}          & \multirow{3}{*}     & \multirow{3}{*}    &32.60    &0.9043  &0.1736     & \multirow{3}{*}      & 31.50     & 0.8422  &0.3160    & \multirow{3}{*}     &30.56       & 0.8181  &0.4039  \\ \midrule
LP-KPN~\cite{cai2019toward}                    & \multirow{6}{*}{\makecell[c]{real-world \\ SISR}}     & \multirow{6}{*}{X2}             & 33.88           &-   &-     &\multirow{6}{*}{X3}        & 32.64       &-  &-       & \multirow{6}{*}{X4}      & 31.58     & - &- \\


CDC~\cite{wei2020component}                     &                           &                        & 34.45     & 0.9104 & 0.1461&     & 33.06       & 0.8762 & 0.2436 &    & 32.42                     & 0.8612 &\textbf{ 0.3005} \\

OR-Net~\cite{li2021learning}               &          &                 & \textbf{34.56}              & 0.910 &- &     & 33.28    & 0.877 &-    & &32.59 & 0.863 &- \\ 
MS2Net~\cite{niu2022ms2net}               & &          &32.74      & 0.9146    &0.1644     &    & -   &-  & - &       &30.07         &0.8544   &0.3799 \\ 
Cria-CL~\cite{shi2022criteria}               &          &                 &-               &-  &- &    & -         &-  &-   &      & 27.85        &0.8112  & 0.3823             \\ 
\textbf{MPF-Net (Ours)}       &                &                 & 34.51     & \textbf{0.9218} & 0.1501    &         & \textbf{33.31}  & \textbf{0.9001}   & 0.2366  &                       &\textbf{32.64}       & \textbf{0.8816} & 0.3104 \\ \hline
\bottomrule
\end{tabular}
}
\end{center}
 \vspace{-10pt}
\end{table*}

\begin{figure*}[!ht]
\centering	
	\includegraphics [width=17 cm]{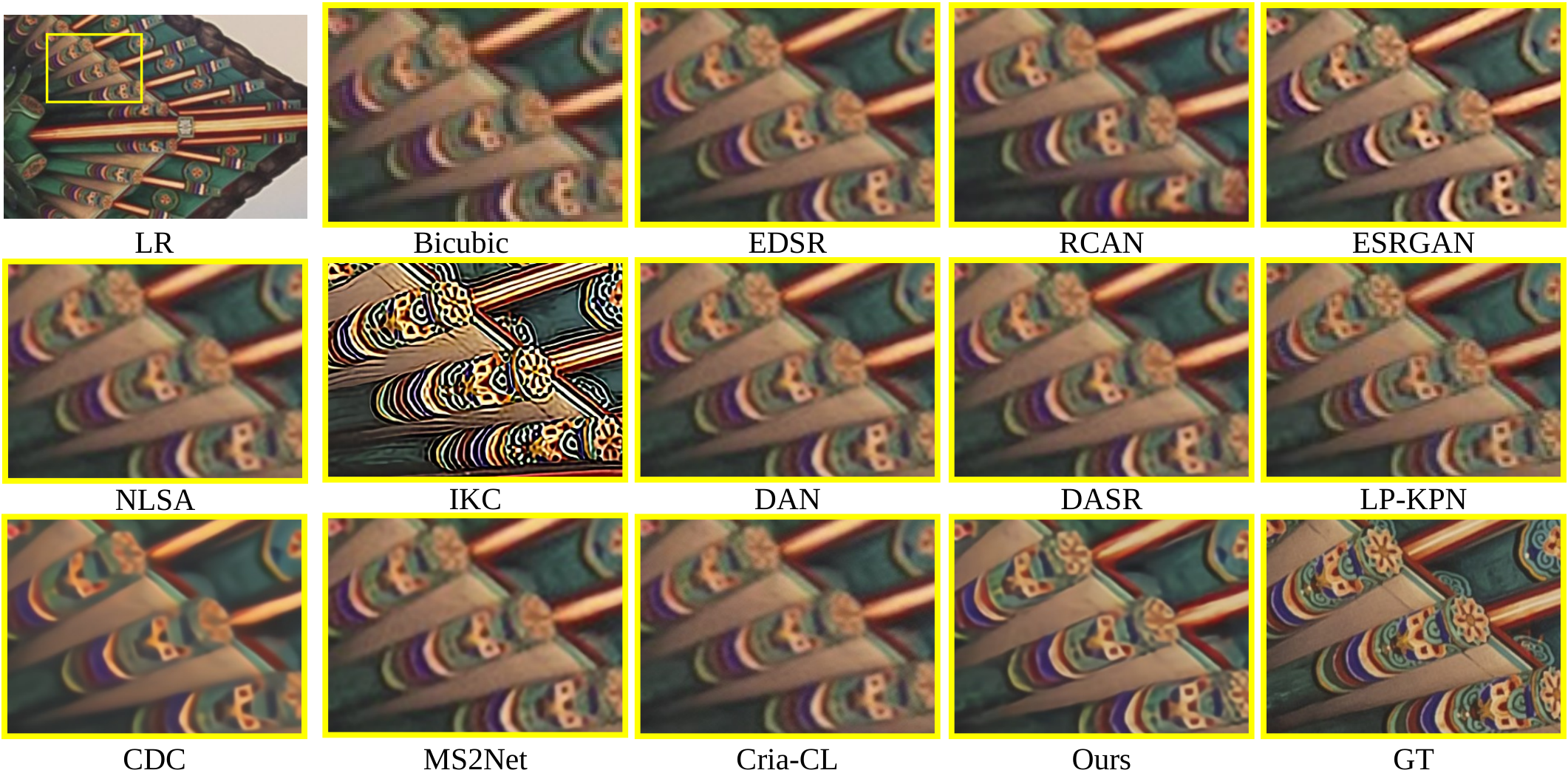}
	\caption{The qualitative comparison of our MPF-Net with the state-of-the-art methods performed on image  `Canon\_017\_LR4'  from RealSR dataset ($\times$4
scale. Zoomed in for a better view.)}
	\label{fig:realsr_x4_1}
\end{figure*}

\begin{figure*}[!ht]
\centering	
	\includegraphics [width=17 cm]{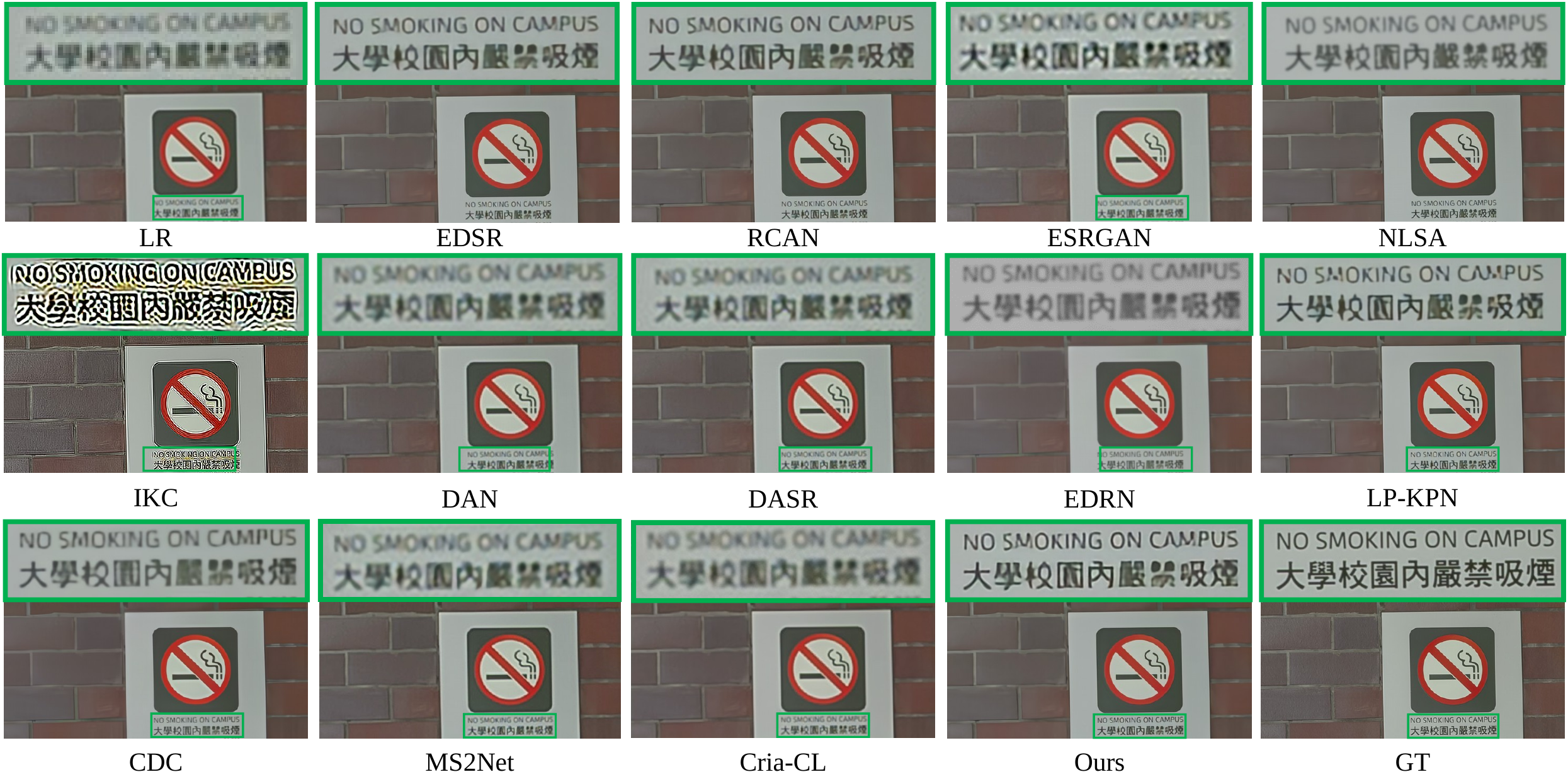}
	\caption{The qualitative comparison of our MPF-Net with the state-of-the-art methods performed on image  `Canon\_001\_LR4'  from RealSR dataset ($\times$4 scale. Zoomed in for a better view.)
    }
	\label{fig:realsr_x4_2}
\end{figure*}

\begin{figure*}[!ht]
\centering	
	\includegraphics [width=18 cm]{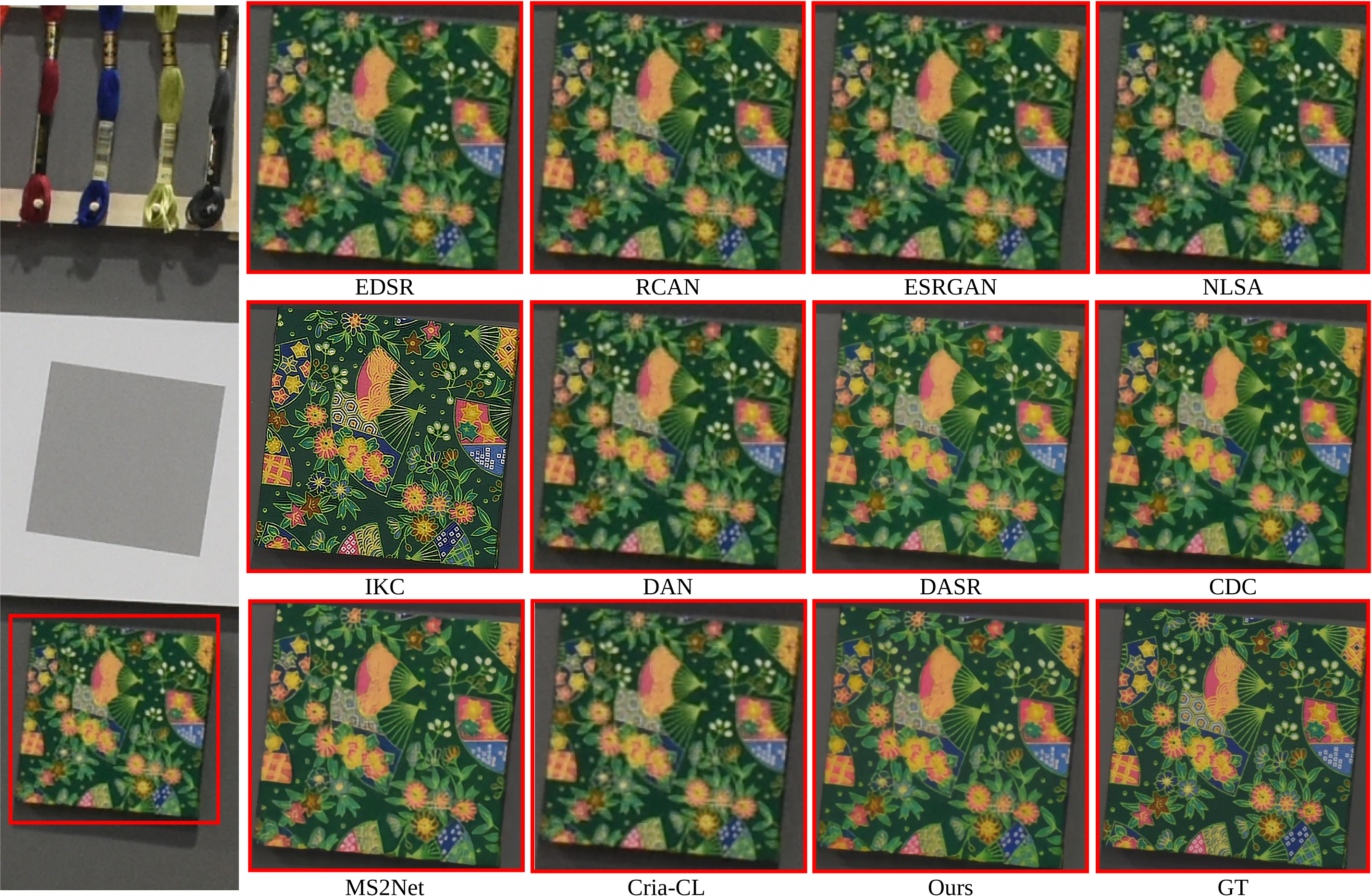}
	\caption{The qualitative comparison of our MPF-Net with the state-of-the-art methods performed on image `DSC\_0988\_x1' from DRealSR dataset ($\times$4
scale, the LR input is cropped from the original one for a better view. All results were zoomed in for a better view.)}
	\label{fig:drealsr}
  \vspace{-10pt}
\end{figure*}

\begin{figure*}[!ht]
\centering	
	\includegraphics [width=18 cm]{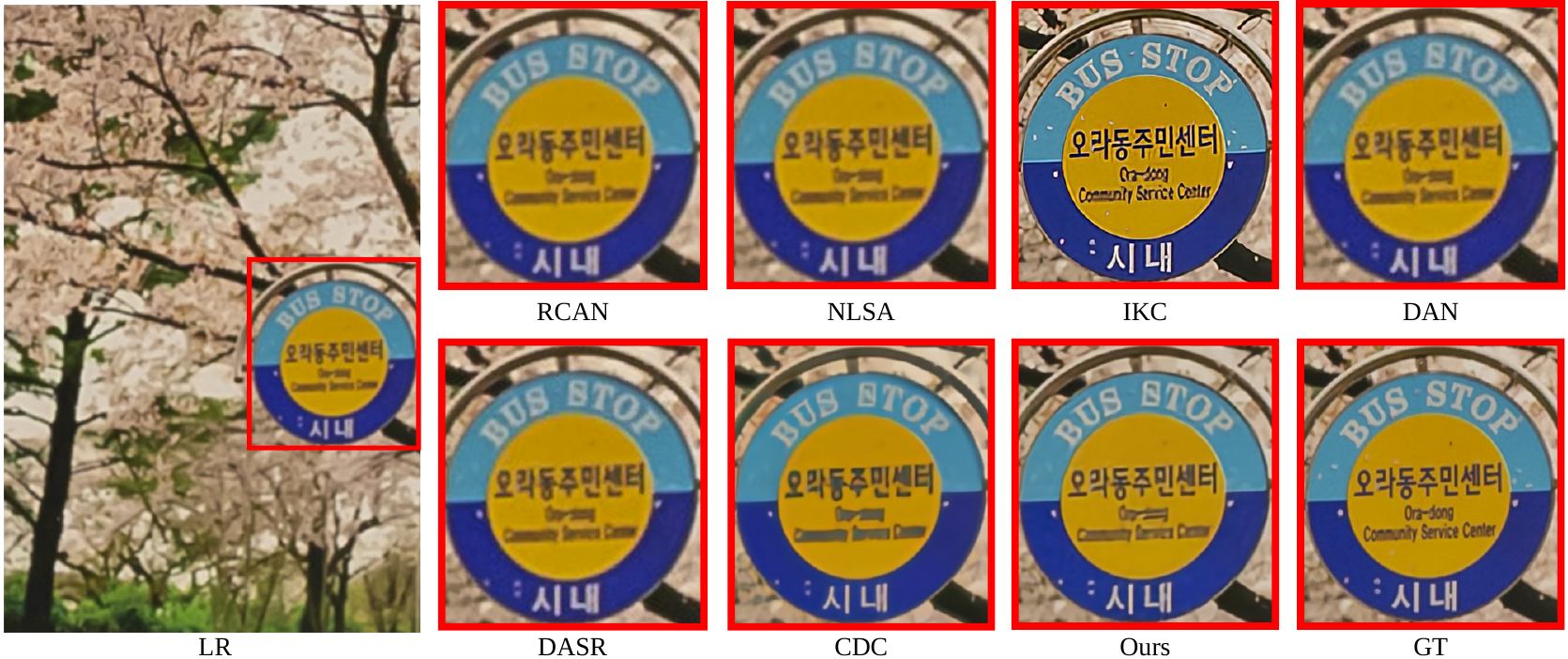}
	\caption{The qualitative comparison of our MPF-Net with the state-of-the-art methods performed on image `Canon\_037\_LR3'  from RealSR dataset ($\times$3
scale. Zoomed in for a better view.)}
	\label{fig:realsr_x3_1}
  \vspace{-10pt}
\end{figure*}

\begin{figure*}[!ht]
\centering	
	\includegraphics [width=18 cm]{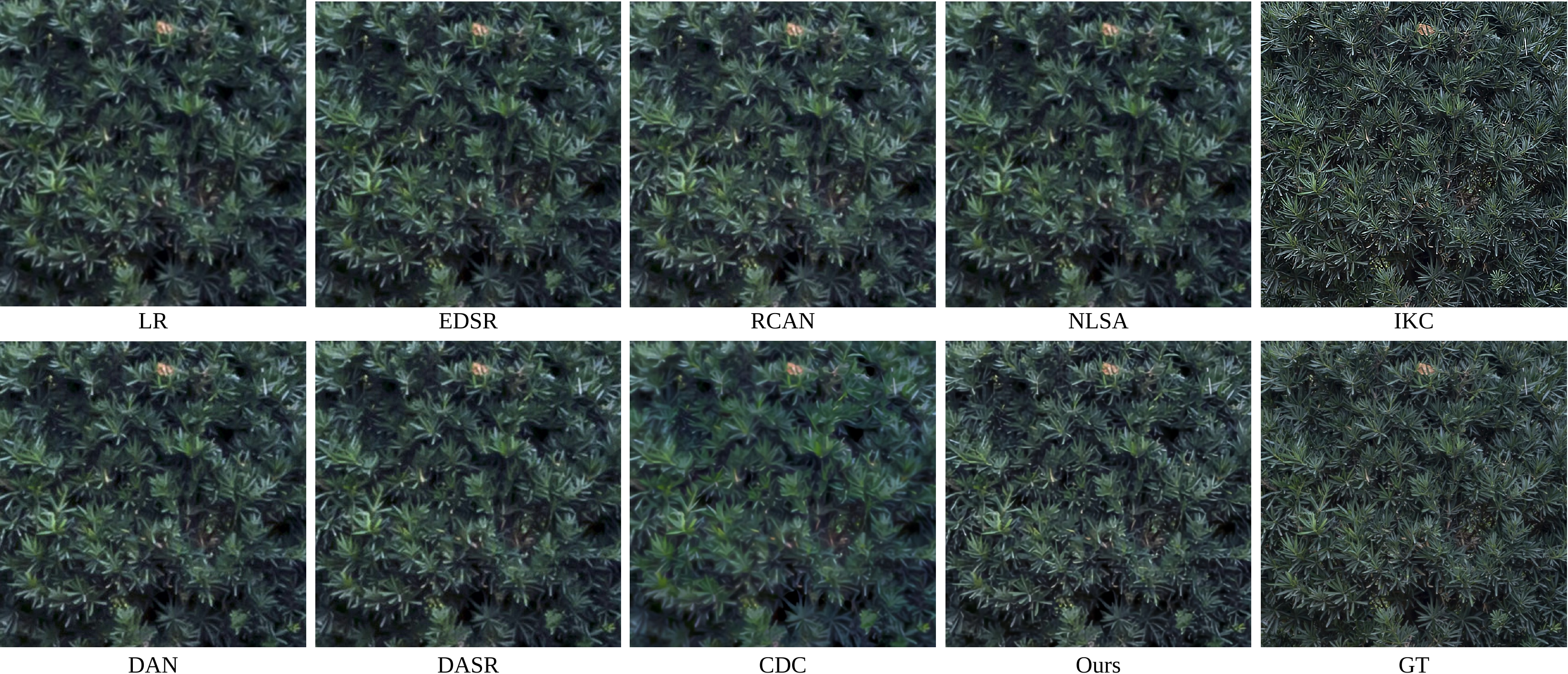}
	\caption{The qualitative comparison of our MPF-Net with the state-of-the-art methods performed on image  `Canon\_001\_LR4'  from RealSR dataset ($\times$3
scale. Zoomed in for a better view.)}
	\label{fig:realsr_x3_2}
  \vspace{-10pt}
\end{figure*}

\textbf{Contrastive Loss.}
We use a modified version of the contrastive loss~\cite{oord2018representation} introduced in~\cite{khosla2020supervised}, which is suitable for a supervised task where there is more than one positive sample. The contrastive loss for the $i$-th image can be formulated as follows:
  
\begin{equation}
    L_{i}=-\frac{1}{M} \sum_{m=1}^{m} \log\frac{\exp(r_{i}^{T}\cdot r_{m}^{+}/\tau  )}{\exp(r_{i}^{T}\cdot r_{m}^{+}/\tau  )+ {\textstyle \sum_{n=1}^{N}(\exp(r_{i}^{T}\cdot r_{n}^{-}/\tau  ))} },
\end{equation}
where ${\tau}$ is the temperature hyper-parameter. ${r_{i}^{T}}$ means the representation of the anchor. ${r_{m}^{+}}$ means the representation of the positive samples, and M is the number of positive samples. N is the number of negative samples, and ${r_{n}^{-}}$ is the corresponding representation. 

In our method, instead of using intermediate features from a pre-trained classification model such as VGG, we use the contrastive discriminator proposed in~\cite{jeong2021training} as our feature embedding network $E$. This network is trained by contrastive learning and has been shown to outperform pre-trained VGG models, as demonstrated in~\cite{wu2021practical}. The architecture of $E$ is based on SNResNet-18, and its first 4 intermediate layers are used to compute the contrastive loss. Thus, the contrastive loss for the $i$-th sample on the $l$-th layer can be defined as follows:

\begin{equation}
\resizebox{1.0\hsize}{!}{$
    L_{i,l}=-\frac{1}{M} \sum_{m=1}^{M} \log\frac{\textstyle \exp(f_{i}^{l}\cdot p_{m}^{l}/\tau  )}{\textstyle \exp(f_{i}^{l}\cdot p_{n}^{l}/\tau )+ {\textstyle \sum_{n=1}^{N}(\exp(f_{i}^{T}\cdot q_{n}^{-}/\tau ))} } $},
\end{equation}
here, the feature representations for the super-resolved image, positive sample, and negative sample are redefined as $f$, $p$, and $q$, respectively. The feature representation of the $i$-th super-resolved image, positive sample, and the negative sample obtained from the $l$-th layer in $E$ are denoted by $f_{i}^{l}$, $p_{m}^{l}$, and $q_{n}^{l}$, respectively. Then, the final contrastive loss $L_{\text{CR}}$ is as follows:
\begin{equation}
    L_{\text{CR}} =\frac{1}{SL} \sum_{i=1}^{S} \sum_{l=1}^{L}L_{i,l},
\end{equation}
where $S$ is the number of images used in the training phase, and $L$  is the number of feature layers from $E$, whose value is set as 4. Therefore, the overall super-resolution loss function Eq.~\ref{eq:loss_total_1} can be further formulated as:

\begin{equation}
    L_{ \text{total}}=L_{\text{recon}}+\beta L_{\text{CR}},
\end{equation}
where $\beta$ is a scaling parameter and we take $\beta$ = 1 as default, which takes the work~\cite{wu2021practical} as reference. We take the $L_{1}$ loss as our $L_{\text{recon}}$.

\begin{figure*}[!ht]
\centering	
	\includegraphics [width=18 cm]{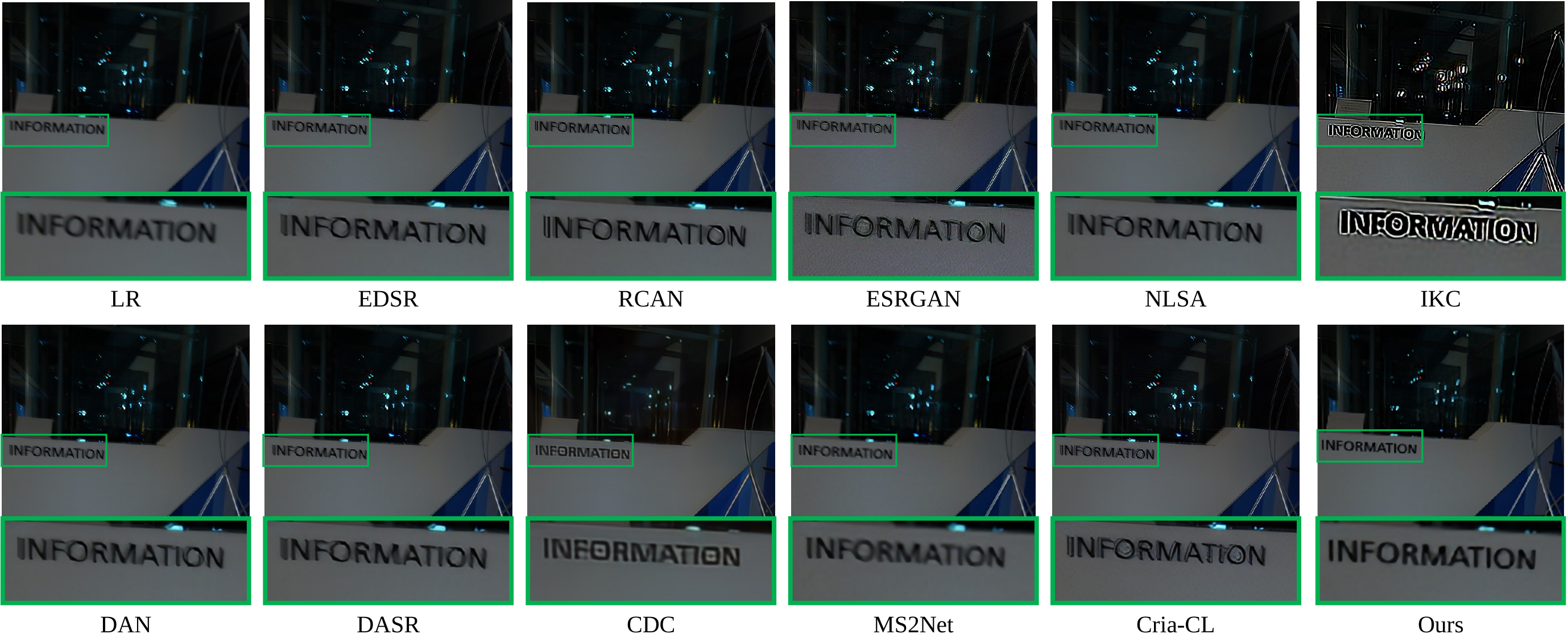}
	\caption{The qualitative comparison of our MPF-Net with the state-of-the-art methods performed on image  `blur\_6'  from scene026 in RealBlur dataset ($\times$4 scale. Zoomed in for a better view.)}
	\label{fig:realBlur}
 \vspace{-16pt}
\end{figure*}

\begin{table*}[!htbp]
\begin{center}
\caption{Quantitative results on the RealBlur dataset ($\times 4$). We compare our MPF-Net to  EDSR, RCAN, ESRGAN, NLSA, IKC, DAN, DASR, CDC, MS2Net, and Cria-CL. We use NIQE as the evaluation metric.}
 \vspace{-10pt}
\label{tab:niqe}
\resizebox{\linewidth}{!}{%
\begin{tabular}{c|c|c|c|c|c|c|c|c|c|c|c}
\toprule
\hline
Methods & EDSR & RCAN & ESRGAN & NLSA & IKC & DAN& DASR& CDC& MS2Net& Cria-CL & \textbf{MPF-Net (Ours)}  \\ \hline
NIQE &5.9765 &5.9946 &7.1626 &5.8261 &5.8038 & 6.3477 & 6.3057 & 7.4639 & 7.1748 &\textbf{ 5.0796} &5.1067 \\ 
\hline
\bottomrule
\end{tabular}
}
\end{center}
 \vspace{-10pt}
\end{table*}

\section{Experiments}

\subsection{ Experiment Setup}
\textbf{Datasets.}  RealSR is a relatively new challenge, and there are only a few datasets available for evaluation. In this work, we evaluate our MPF-Net on three datasets: RealSR~\cite{cai2019toward}, DRealSR~\cite{wei2020component}, and RealBlur~\cite{rim2020real}. RealSR~\cite{cai2019toward} comprises 3,147 images collected from 559 scenes captured by Canon 5D3 and Nikon D810 devices. The dataset has three scales (2$\times$, 3$\times$, and 4$\times$), with 400 image pairs for training and 100 for testing in each scale. The training samples are collected from 459 scenes, and the testing images are from 100 scenes. DRealSR~\cite{wei2020component} consists of 35,065, 26,118, and 30,502 image patches for scales of $\times2$, $\times3$, and $\times4$, respectively, in the training dataset. The testing dataset includes 83, 84, and 93 images for $\times2$, $\times3$, and $\times4$, respectively. The image sizes for patches of scales $\times$2, $\times$3, and $\times$4 are 380$\times$380, 272$\times$272, and 192$\times$192, respectively.
RealBlur~\cite{rim2020real} is captured under real-world conditions and consists of two subsets: (1) RealBlur\_J, formed with camera JPEG outputs, and (2) RealBlur\_R, generated offline by applying white balance, demosaicking, and denoising operations to the RAW images.

\textbf{Implementation details.} Our method is implemented using PyTorch 1.12.0 and trained on one NVIDIA TITAN RTX GPU. We use the Adam optimizer with exponential decay rates $\beta_{1}$ and $\beta_{2}$ set to 0.9 and 0.999, respectively. We first train the SR network by optimizing ${L_{\text{recon}}}$ for 200000 iterations with an initial learning rate of ${1\times10^{-4}}$ and a batch size of 32. Next, we train the entire network for another 300000 iterations using ${L_{\text{total}}}$. To adjust the learning rate, we use the cosine annealing strategy~\cite{wu2021contrastive} and decay the learning rate using a cosine schedule. We perform data augmentation by randomly flipping, rotating, and cropping the training LR image to 48$\times$48 for scales 2, 3, and 4.

All quantitative evaluations use PSNR, SSIM, LPIPS, and NIQE as error measures and are conducted on the luminance channel, as is commonly done in the existing literature. PSNR is an objective evaluation method that measures pixel-level error-based differences. SSIM is a subjective evaluation method that incorporates the visual perceptual characteristics of the human eye, including information about the image's brightness, contrast, and internal structure. LPIPS~\cite{zhang2018unreasonable} is a reference-based image quality evaluation metric that computes the perceptual similarity between the ground truth and the SR image. NIQE~\cite{mittal2012making} is a no-reference image quality score that is based on a collection of statistical features designed to capture the quality of natural scene statistics in space-domain images.

\vspace{-10pt}
\subsection{ Comparison with State-of-the-art Methods}

We evaluated our MPF-Net against various state-of-the-art SISR models, including traditional learning-based models such as Bicubic, EDSR~\cite{lim2017enhanced}, RCAN~\cite{zhang2018image}, ESRGAN~\cite{wang2018esrgan}, NLSA~\cite{mei2021image}, kernel-based methods likeIKC~\cite{gu2019blind}, DAN~\cite{huang2020unfolding}, DASR~\cite{wang2021unsupervised}, as well as real-world SISR models such as LP-KPN~\cite{cai2019toward}, CDC~\cite{wei2020component}, NLSA~\cite{mei2021image}, OR-Net~\cite{li2021learning}, MS2Net~\cite{niu2022ms2net}, Cria-CL~\cite{shi2022criteria}. The results were obtained using the codes provided by the authors or the corresponding papers. Our MPF-Net outperformed all other models in terms of PSNR and SSIM, as shown in Tab.\ref{tab:realsr} and Tab.\ref{tab:drealsr}.

We found that traditional learning-based SISR models are limited to solving synthetic degradation and cannot be generalized to other cases. Although RCAN and NLSA adopted attention and non-local technologies, they still struggled with real-world images and required more computational resources. Kernel-based SISR methods aim to evaluate the degraded kernel for the input LR image and design an SR strategy based on the evaluated prior. However, if the evaluated degraded kernel does not match the true kernel, the trained model will collapse. Moreover, the existing real-world SISR models, such as LP-KPN, CDC, and OR-Net, reconstruct image details relying on shallow features obtained from a single receptive field. They ignore the features obtained from different receptive field convolutions, which can provide adaptive multi-scale reception ability and significantly enrich the network's reception scales and perceptual ability.

We present a qualitative comparison of our MPF-Net with the above-mentioned methods in terms of visual results, as shown in Fig.\ref{fig:realsr_x4_1}, Fig.\ref{fig:realsr_x4_2}, and Fig.\ref{fig:drealsr}. We observe that traditional SISR methods such as bicubic, EDSR, RCAN, and ESRGAN fail to restore some corrupted details in the real LR images, such as the alphabet text in the figures. Due to the complexity of the degradation process in real-world images, it is difficult to estimate and quantify the degradation kernel accurately. Therefore, kernel-based SISR methods do not generalize well on real-world images. In comparison, our MPF-Net can better restore textures and details and generate results closer to the ground truth images than real-world SISR methods. We also provide visual results under $3\times$ scale in Fig.\ref{fig:realsr_x3_1} and Fig.~\ref{fig:realsr_x3_2}, which further demonstrate the superior performance of our MPF-Net in generating high-quality visual results.

\textbf{RealBlur dataset.} 
To demonstrate the ability of our MPF-Net to generalize to unknown degradation conditions, we applied our method to the RealBlur dataset~\cite{rim2020real}, which contains motion-blurred images. We selected the first 16 scenes from RealBlur\_J, resulting in a total of 336 images, and directly down-sampled the blurry images to generate the blur LR images. Fig.\ref{fig:realBlur} presents visual comparisons for 4$\times$ super-resolution. The methods EDSR\cite{lim2017enhanced}, RCAN~\cite{zhang2018image}, ESRGAN~\cite{wang2018esrgan}, and NLSA~\cite{mei2021image} were not designed for motion-blurred images and, thus, show almost no improvement over the input image. Kernel-based methods heavily rely on the estimated kernels, leading to unsatisfactory performance. Compared to current real-world image super-resolution methods, our MPF-Net produces more natural results with clearer texts that can be easily recognized. Moreover, considering that the RealBlur dataset is a sequence set without one-to-one correspondence with ground truth, we use the NOIQ metric as the evaluation standard to compare our method with others. As shown in Tab.~\ref{tab:niqe}, our method also has a slight advantage.

\subsection{Analysis on Complexity} 

As shown in Tab. \ref{tab:param_flop_time}, we report the results of the computation and complexity comparison of different methods. Among them, EDSR \cite{lim2017enhanced} has the least computation, requiring only 99.51G FLOPs and 1.52M parameters for a patch size of $224\times 224$. However, its performance is relatively poor, with PSNR, SSIM, and LPIPS of 29.09, 0.8270, and 0.2779, respectively. As performance improves, both the number of parameters in models and their demands on computing resources are increasing. For instance, RCAN has 15.59M parameters and requires 799.33G FLOPs, while ESRGAN has 16.70M parameters and requires 899.50G FLOPs. On the other hand, NLSA \cite{mei2021image} has the highest computation, with 2574.91G FLOPs and 44.16M parameters, indicating a higher demand for computational resources for non-local techniques. In comparison, kernel-based methods have relatively fewer model parameters. Specifically, IKC, DAN, and DASR have 4.24M, 4.32M, and 5.95M parameters, respectively. However, due to the need for iterative estimation of more accurate kernels in DAN, its computational requirements are higher, specifically 976.32G FLOPs.

Compared with these existing real-world super-resolution methods, our MPF-Net achieves the best super-resolution results on the given RealSR test dataset with only 5.95M parameters and 327.15G FLOPs of computational resources. In contrast, CDC, MS2Net, and Cria-CL require 39.92M, 13.96M, and 17.29M parameters, as well as 706.96G, 446.55G, and 1370.71G FLOPs, respectively. Although our method's average inference time is 0.9743 seconds for a single image, slightly longer than NLSA's 0.9697 seconds, our method has achieved relatively superior results in quantitative and complexity analysis compared to the current SOTA real-world super-resolution methods.

\begin{table*}[!htbp]
\begin{center}
\caption{Computation and complexity of different methods. The size of the parameter is measured in MB, while FLOPs are measured in Gmac for 224x224 images. The running time is measured in seconds per image, averaged across 100 images in the RealSR dataset. The top-performing and second-best results are both highlighted and underlined.}
\label{tab:param_flop_time}
\begin{tabular}{c|c|c|c|c|c|c|c}
\toprule
\hline
Methods & Category & Params (M)$\downarrow$ & FLOPs (G)$\downarrow$ & Time (s)$\downarrow$ & PSNR & SSIM & LPIPS \\ \hline
Bicubic & \multirow{5}{*}{ \makecell[c]{ vanila \\ SISR}} & 0.00     & 0.62    & 0.0002  &27.24 &0.7635 &0.4764 \\ 
EDSR    & & \textbf{1.52}     & \textbf{99.51}    & 0.0474  &29.09 &0.8270 &0.2779  \\ 
RCAN    & & 15.59    & 799.33   & 0.3471  &29.21 &0.8237 &0.2868 \\ 
ESRGAN  & & 16.70    & 899.50   & 0.5124  &29.15 &0.8263 &0.2793 \\ 
NLSA    & & 44.16    & 2574.91  & 0.9697  &29.23 &0.8281 &0.2651 \\\midrule
IKC     & \multirow{3}{*}{ \makecell[c]{ kernel-based \\ SISR}} & 4.24     & 245.92   & 0.5240  &16.81 &0.5277 &0.3824 \\
DAN     & & 4.32     & 967.32   & 0.6558  &27.80 &0.7882 &0.4114 \\ 
DASR    & & 5.95     & 159.86   & 0.0519  &27.79 &0.7874 &0.4076  \\ \midrule
CDC     & \multirow{4}{*}{ \makecell[c]{ real-world \\ SISR}} & 39.92    & 706.96   &0.1709       &29.24 &0.8265 &0.2781  \\  
MS2Net  & & 13.96     & 446.55   &\textbf{0.0394}       &27.33 &0.7869 &0.3565  \\ 
Cria-CL & & 17.29    &1370.71   &0.2378  &25.83 &0.7324 &-  \\ 
\textbf{MPF-Net (Ours)}     & & 5.95     & 327.15   & 0.9743 &\textbf{29.50} &\textbf{0.8288} &\textbf{0.2664}  \\ \hline
\bottomrule
\end{tabular}
\end{center}
 \vspace{-10pt}
\end{table*}

\subsection{Ablation Study}
To study the efficiency of our MPF-Net, we design and conduct the following ablation experiments :
 
\begin{figure}[!ht]
\centering	
	\includegraphics [width=9 cm]{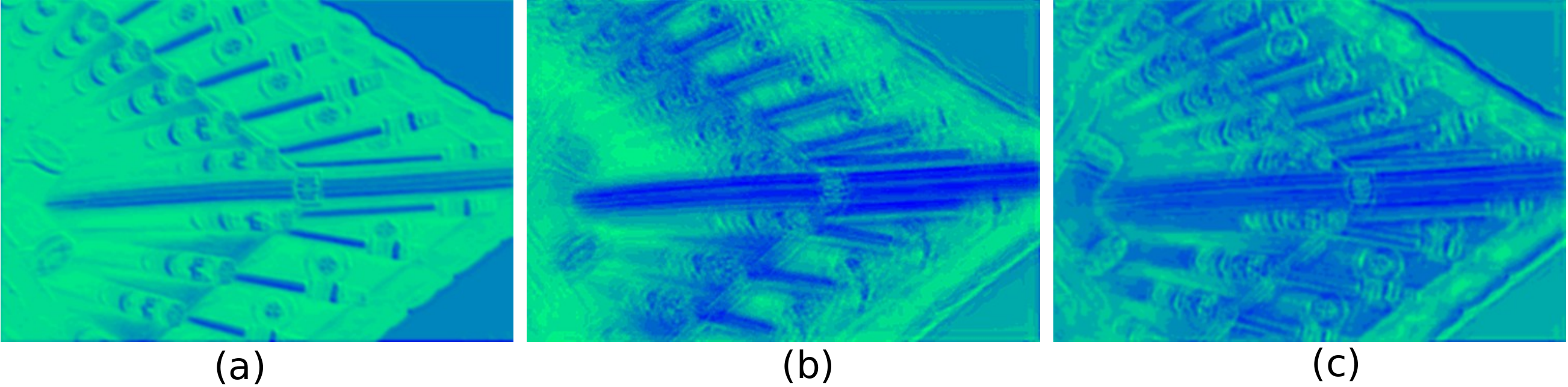}
	\caption{The visualization of the extracted feature map from the different convolutional operations. (a) Feature map from 5 same vanilla convolutions; (b) Feature map from 5 dilated convolutions (dilation = 1,2,3,6,9) ; (c) Feature map from our MPEF. (For a better view, we add all output feature maps and mean them.)
 }
	\label{fig:heatmap}
  \vspace{-18pt}
\end{figure}

\textbf{Effectiveness of the proposed MPFE and CPB in our MPF-Net.} To test the effectiveness of our MPFE and CPB, we replaced them with vanilla convolution operations and non-linear mapping. The experimental settings were as follows:

\textbf{1) Baseline.}  we employ an MPFE unit without dilated convolution, comprising five vanilla convolutions with a 3 $\times$ 3 kernel size. We then use similar CPB blocks, but without the cross-fusion from up-to-down and down-to-up connections, to perform nonlinear mapping;

\textbf{2) Baseline + CPB.} Expanding on 1), we utilize our proposed CPB block that incorporates cross-fusion with both up-to-down and down-to-up connections to perform nonlinear mapping;

\textbf{3) Baseline + MPFE.} Building upon 1), we enhance our feature extraction by utilizing our proposed MPFE, which consists of two vanilla convolution layers and three different scale dilated convolution layers to extract shallow features.

\textbf{4) Baseline + MPFE + CPB.} The model used in our paper.

\begin{table}[!t]
\centering	
\caption{Ablation experiments conducted on RealSR dataset to study the effectiveness of the proposed MPFE unit and CPB block designed in our MPF-Net.}
\label{tab:mpfe_cpb}
\begin{tabular}{c|c|c|c|c}
\toprule
\hline
Case & MPFE & CPB & PSNR & SSIM \\
\midrule
baseline &\xmark             & \xmark      &29.12      & 0.8212     \\ 
baseline + CPB &\xmark       & \cmark      &29.19      &0.8221       \\
baseline + MPFE &\cmark      & \xmark         &29.37      &0.8236     \\
baseline + MPFE + CPB &\cmark    & \cmark     &29.50      &0.8288      \\\hline
\bottomrule
\end{tabular}
\end{table}

In Section~\ref{sec:MPEF}, we have explained how our MPFE can generate multi-perception features that encompass more detailed information, which is highly significant for achieving image super-resolution. Tab.\ref{tab:mpfe_cpb} demonstrates that our baseline network's performance can be enhanced through MPFE. We observe an increase in PSNR by 0.25dB and 0.31dB from the baseline to baseline + MPFE and from baseline + CPB to baseline + MPFE + CPB, respectively. To further illustrate the effectiveness of our MPFE, we visualize the features obtained through conventional feature extraction methods and our MPFE. Fig.~\ref{fig:heatmap} depicts the feature maps: (a) shows the output from 5 conventional convolutions, (b) represents the output from 5 dilated convolutions, and (c) illustrates the output from our MPFE. We observe that the feature map in (a) is too smooth and lacks detail, whereas the feature map in (b) contains more detail but is somewhat blurred. The feature map obtained through our MPFE is both smooth and detailed.

\textbf{Evaluation on MPFE.} Here, we conducted further experiments using different settings to examine the impact of multi-perspective features. Tab.\ref{tab:mpfe} displays the results. The first line, with zero vanilla conv and five dilated convs, indicates that the MPFE consists of three convolutions with dilation=3, one with dilation=6, and one with dilation=9. The second line, with one vanilla conv and four dilated convs, means that the MPFE includes one vanilla convolution with a $3\times3$ kernel size, three convolutions with dilation=3, one with dilation=6, and one with dilation=9. The third line, with two vanilla convs and three dilated convs, indicates that the MPFE includes two vanilla convolutions with kernel sizes of $1\times1$ and $3\times3$, respectively, as well as one convolution with dilation=3, one with dilation=6, and one with dilation=9. The fourth line, with three vanilla convs and two dilated convs, means that the MPFE includes three vanilla convolutions with one $1\times1$ convolution and two $3\times3$ convolutions, as well as one convolution with dilation=6 and one with dilation=9. The fifth line, with four vanilla convs and one dilated conv, indicates that the MPFE includes four vanilla convolutions with one $1\times1$ convolution and three $3\times3$ convolutions, as well as one convolution with dilation=6. Finally, the sixth line, with five vanilla convs and zero dilated convs, means that the MPFE includes five vanilla convolutions with one $1\times1$ convolution and four $3\times3$ convolutions. Due to the convolution kernel's size and different receptive fields, there are numerous combinations. Therefore, we chose a few typical settings to display in the table.

\begin{figure}[!ht]
\centering	
	\includegraphics [width=9 cm]{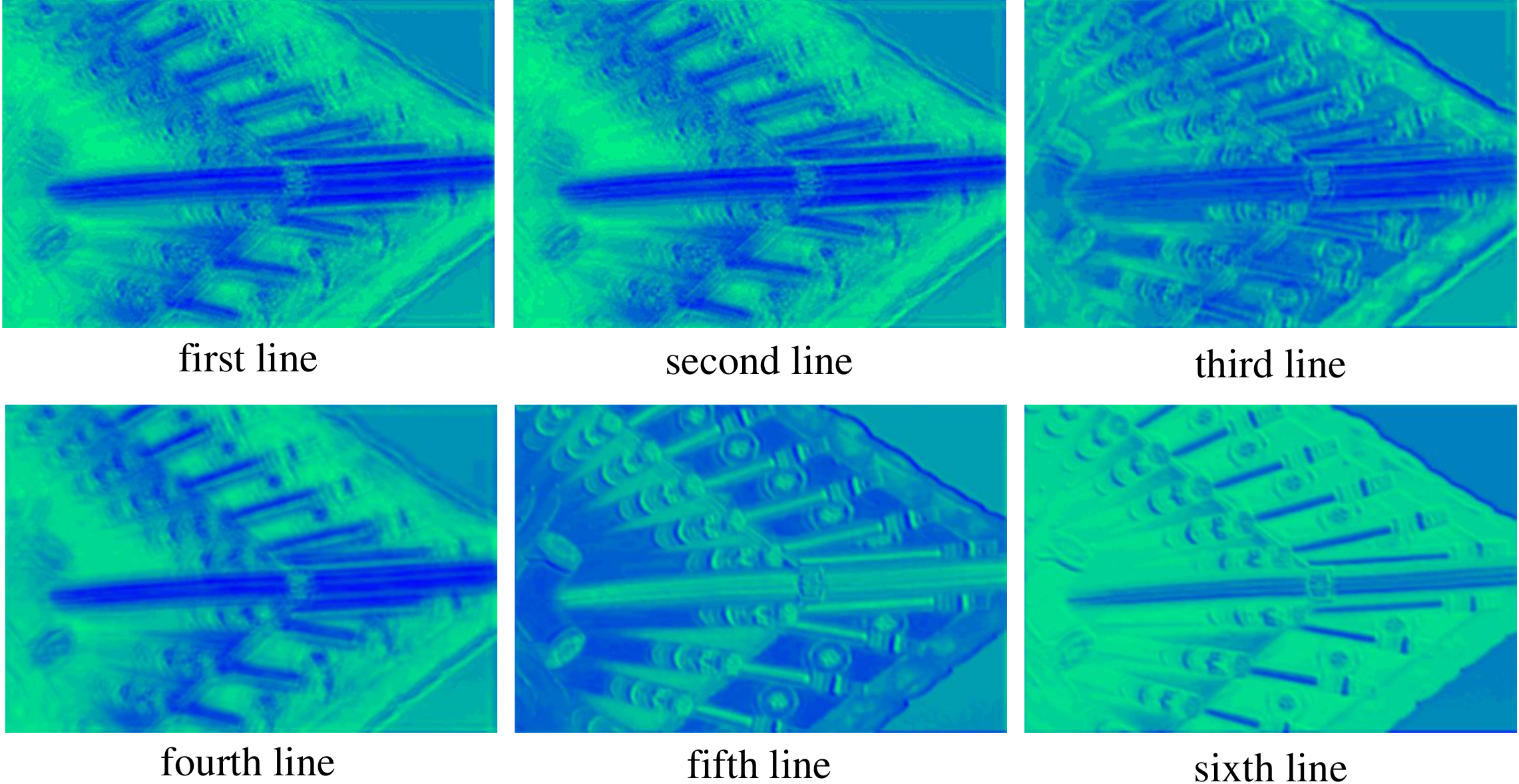}
	\caption{The visualization of the extracted feature map from the different MPEF with different settings.}
	\label{fig:multi-perception}
\end{figure}

The results from the second line of the experiments show that smooth feature extraction from vanilla convolutions is also essential for improving performance compared to the first line. In contrast, the results from the second, third, fourth, and fifth lines indicate that multi-perception features are critical for boosting performance compared to the final line. Moreover, we provide visualizations of the extracted features from the various MPFE settings. As displayed in Fig.~\ref{fig:multi-perception}, the visualizations show that the more vanilla convolutions used, the smoother the extracted features but with less detail. On the other hand, using more dilation convolutions results in sharper extracted features but with overlapping details.

\begin{table}[!t]
\centering	
\caption{Ablation experiments of MPFE module with different numbers of standard convolution and dilated convolution layers, conducted on RealSR. $m+n=5$ }\label{tab:mpfe}
\begin{tabular}{c|c|c|c}
\toprule
\hline
vanilla conv ($m$) & dilated conv ($n$) & PSNR & SSIM \\
\midrule
0             & 5           &29.126 &0.8189 \\ 
1             & 4           &29.225 &0.8208       \\
2             & 3           &29.502 &0.8288       \\
3             & 2           &29.434 &0.8259       \\
4             & 1           &29.371 &0.8245       \\
5             & 0           &29.191 &0.8221       \\
\hline
\bottomrule
\end{tabular}
\end{table}

\begin{table}[!t]
\centering	
\caption{Ablation study results. $L_{PCR}$ and $L_{VCR}$ denote our proposed CR and that utilizing the pre-trained VGG model, respectively. $P_{M}$ is positive set and  $N_{N}$ represents the negative one. LR, GT, Rand, and Gen utilize only LR input, only ground truth image,  randomly selected other instances in the same batch size, and generated samples, respectively.}
\label{tab:cr}
\scalebox{0.9}{
\begin{tabular}{c c c c c c |cc}
\toprule
\hline
Config & $L_{1}$ &$L_{PCR}$ &$L_{VCR}$ &$P_{M}$  &$N_{N}$ & PSNR & SSIM \\
\midrule
 1 &\cmark   &      &        &        &         &29.373       &0.8236    \\  
 2 &\cmark   &      &\cmark  &Gen+GT  &Gen+LR   &29.416       &0.8254    \\
 3 &\cmark   &\cmark&        &Gen     &Gen      &29.445       &0.8261    \\
 4 &\cmark   &\cmark&        &Gen+GT  &Ran      &29.431       &0.8251    \\
 5 &\cmark   &\cmark&        &Gen+GT  &LR       &29.469       &0.8267    \\
 6 &\cmark   &\cmark&        &GT      & Gen+LR  &29.455       &0.8261    \\  
 7 &\cmark   &\cmark&        &Gen+GT  & Gen+LR  &29.502       &0.8288    \\\hline  
\bottomrule
\end{tabular}
}
\end{table}

\textbf{Evaluation on Contrastive Regularization.}  

The effectiveness of our proposed Contrastive Regularization (CR) term, as described in Section~\ref{sec:CR}, is demonstrated through ablation studies conducted on the RealSR dataset, and the results are presented in Tab.\ref{tab:cr}. The metrics used to evaluate the performance are PSNR and SSIM. Our proposed approach (\textbf{Config. 7}) outperforms all other configurations, achieving the best results on both metrics. Comparing the results of~\textbf{Config. 5},~\textbf{6}, and \textbf{7} highlights the effectiveness of using generated negative and positive samples. Additionally, it can be observed that the use of low-resolution (LR) inputs in~\textbf{Config. 4} and~\textbf{5} results in better performance than randomly selected samples of the same batch size. This finding is consistent with the claim made in~\cite{wu2021practical}. The comparison between~\textbf{Config. 2} and~\textbf{7} demonstrates that our proposed $L_{PCR}$ works well in RGB space and remains comparable to the $L_{VCR}$ built upon the VGG pre-trained model, as stated in~\cite{wu2021practical}. Finally, our method shows superior performance over~\textbf{Config. 3} by incorporating both the LR and ground truth (GT) into CR.

\section{Conclusion}

We present the~\textbf{ MPF-Net}, a novel approach for real-world single-image super-resolution that achieves state-of-the-art results. Our method leverages multi-perception features to extract more local and global information from the input image, which leads to better reconstruction of details. The MPF-Net comprises three modules: the multi-perception feature extraction unit (MPFE), Cross-Perceived Block (CPB), and Contrastive Regularization (CR). The CPB allows for complementary use of the obtained local and global information, improving the information in each perceptual domain. Additionally, the CR loss is built upon specially generated positive and negative samples, which better pushes the reconstructed image towards the clear one in the representation space. We conduct extensive experiments on multiple datasets, including RealSR, DRealSR, and RealBlur, in various settings. Our MPF-Net consistently outperforms existing state-of-the-art methods by a significant margin in both subjective and objective evaluations.

Regarding the limitations of our proposed method, we acknowledge that although it outperforms many state-of-the-art (SOTA) methods on many real-world images, such as RealSR~\cite{cai2019toward}, DRealSR~\cite{wei2020component}, and RealBlur~\cite{rim2020real}, the results obtained by our method still lack some details compared with the ground truth. Additionally, the inference time is not as efficient as some current advanced methods (as shown in Tab.~\ref{tab:param_flop_time}). To address these limitations, we plan to continue improving the performance and efficiency of our method in future work to enhance its generalizability by applying some other technologies, such as sparse-coding~\cite{yang2010image,niu2023gran}.

\ifCLASSOPTIONcaptionsoff
  \newpage
\fi


\bibliographystyle{IEEEtran}
\bibliography{mybibfile.bib}

%

\vspace{-20pt}
\begin{IEEEbiography}
[{\includegraphics[width=1in,height=1.25in,clip,keepaspectratio]{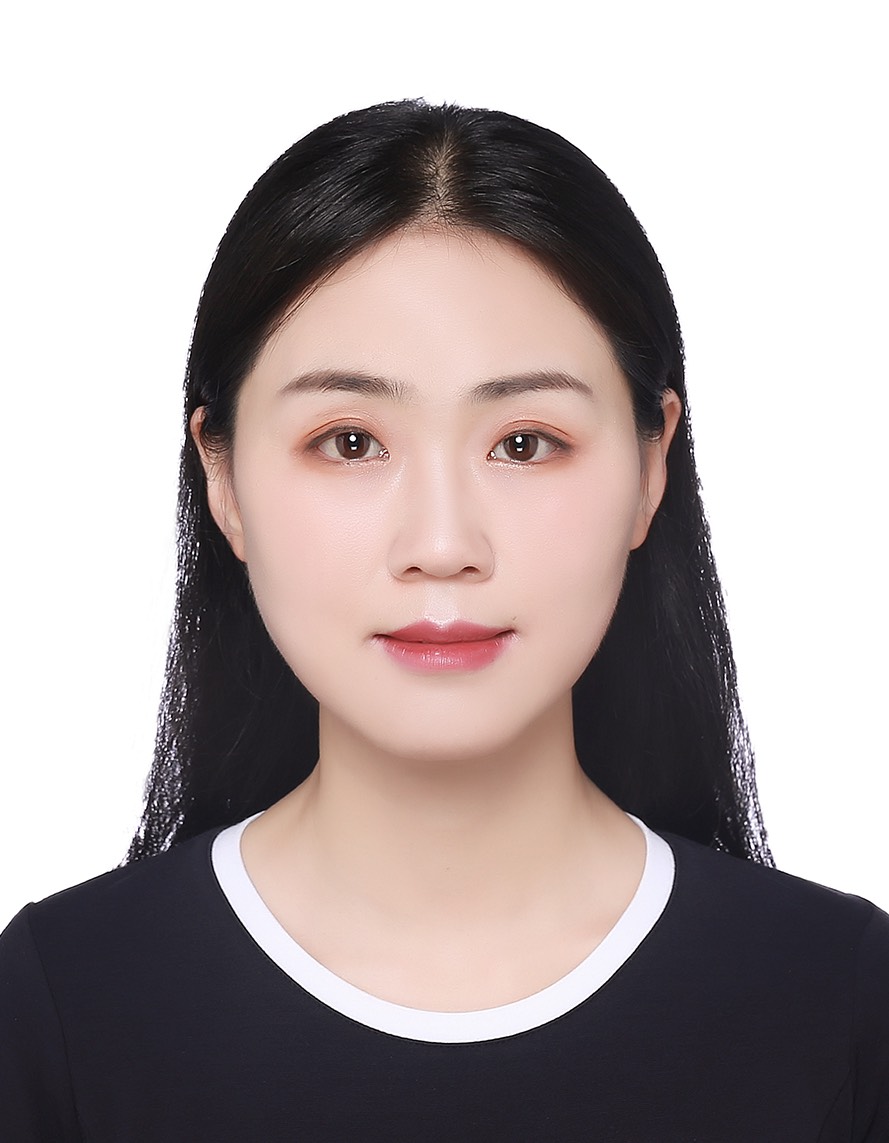}}] {Axi Niu} received her B.S. and M.S. degrees from the Henan  University, Kaifeng, China, in 2014 and 2017. She is currently pursuing the Ph.D degree with the School of Computer Science, Northwestern Polytechnical University, Xi’an, China. Her research interests include image processing and computer vision.
\end{IEEEbiography}
\vspace{-30pt}

\begin{IEEEbiography}
[{\includegraphics[width=1in,height=1.25in,clip,keepaspectratio]{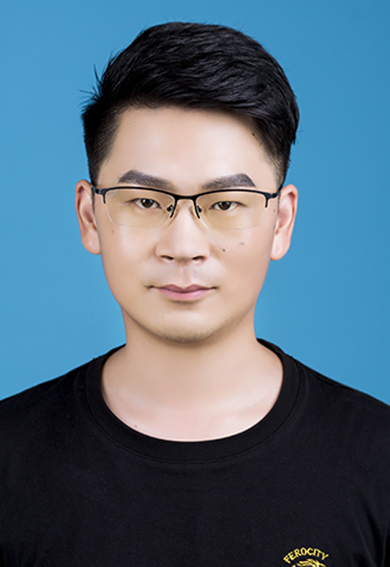}}] {Kang Zhang} received his B.S. degree from Harbin Institute of Technology, 2020. He is currently pursuing the Ph.D degree at Korea Advanced Institute of Science \& Technology. His research work focuses on Deep Learning, Self-Supervised Learning, and Adversarial Machine Learning. 
\end{IEEEbiography}
\vspace{-30pt}

\begin{IEEEbiography}
[{\includegraphics[width=1in,height=1.25in,clip,keepaspectratio]{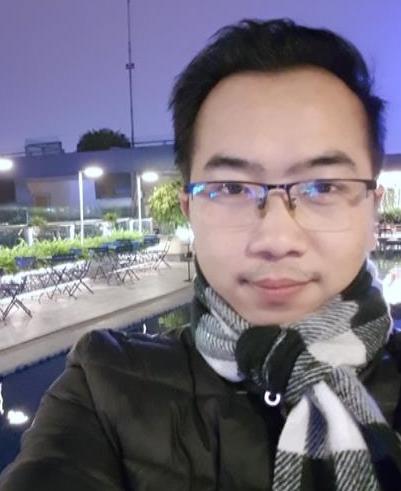}}] { Pham Xuan Trung} received his B.S. degree in the School of Electronics and Telecommunications (SET) at Hanoi University of Science and Technology (HUST) in 2014. He is currently working toward his Ph.D. at KAIST under the supervision of Prof. Chang D. Yoo. His doctoral
research interests include Speech Processing, SelfSupervised Learning, and Computer Vision.
\end{IEEEbiography}
\vspace{-30pt}

\vspace{-30pt}
\begin{IEEEbiography}
[{\includegraphics[width=1in,height=1.25in,clip,keepaspectratio]{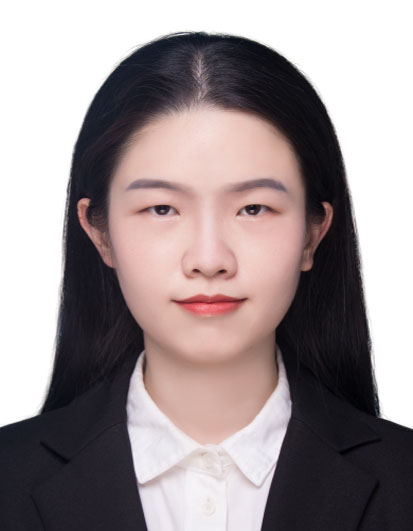}}]
{Pei Wang} received her B.S. degree from the Shaanxi Normal University, Xi’an, China,
in 2016. She is currently pursuing the Ph.D degree with the School of Computer
Science, Northwestern Polytechnical University, Xi’an, China. Her research interests
include image deblurring and computer vision.
\end{IEEEbiography}
\vspace{-30pt}

\vspace{-30pt}
\begin{IEEEbiography}
[{\includegraphics[width=1in,height=1.25in,clip,keepaspectratio]{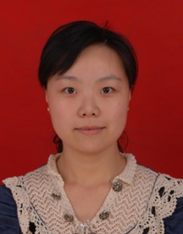}}]
{Jinqiu Sun} received her B.S., M.S. and Ph.D. degrees from Northwestern Polytechnical University in 1999, 2004 and 2005, respectively. She is presently a Professor of School of astronomy, Northwestern Polytechnical University. Her research work focuses on signal and image processing, computer vision and pattern recognition.
\end{IEEEbiography}
\vspace{-30pt}


\vspace{-30pt}
\begin{IEEEbiography}
[{\includegraphics[width=1in,height=1.25in,clip,keepaspectratio]{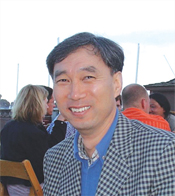}}]
{In So Kweon} received the B.S. and the M.S. degrees in Mechanical Design and Production Engineering from Seoul National University, Korea, in 1981 and 1983, respectively, and the Ph.D. degree in Robotics from the Robotics Institute at Carnegie Mellon University in 1990.  He is currently a Professor of electrical engineering (EE) and the director for the National Core Research Center – P3 DigiCar Center at KAIST. He served as the department head of Automation and Design Engineering (ADE) at KAIST in 1995-1998. His research interests include computer vision and robotics. He has co-authored several books, including "Metric Invariants for Camera Calibration," and more than 300 technical papers. He served as a Founding Associate-Editor-in-Chief for “International Journal of Computer Vision and Applications”, and has been an Editorial Board Member for “International Journal of Computer Vision” since 2005. He is  a member of many computer vision and robotics conference program committees and has been a program co-chair for several conferences and workshops. Most recently, he is a general co-chair of the 2012 Asian Conference on Computer Vision (ACCV) Conference. He received several awards from international conferences, including “The Best Student Paper Runnerup Award in the IEEE-CVPR’2009” and “The Student Paper Award in the ICCAS’2008”. He also earned several honors at KAIST, including the 2002 Best Teaching Award in EE. In 2001, he received the KAIST Research Award. He is a member of KROS, ICROS, and IEEE.
\end{IEEEbiography}
\vspace{-30pt}

\vspace{-30pt}
\begin{IEEEbiography}
[{\includegraphics[width=1in,height=1.25in,clip,keepaspectratio]{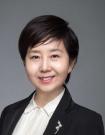}}]
{Yanning Zhang} received her B.S. degree from Dalian University of Science and
Engineering in 1988, M.S. and Ph.D. Degree from Northwestern Polytechnical University in 1993 and 1996, respectively. She is presently a Professor of School of
Computer Science and Technology, Northwestern Polytechnical University. She is
also the organization chair of ACCV2009 and the publicity chair of ICME2012. Her
research work focuses on signal and image processing, computer vision and pattern recognition. She has published over 200 papers in these fields, including the
ICCV2011 best student paper. She is a member of IEEE.
\end{IEEEbiography}
\vspace{-30pt}

\end{document}